\pdfoutput=1
\documentclass{article}

\PassOptionsToPackage{table}{xcolor}

\usepackage{iclr2027_conference,times}

\usepackage[utf8]{inputenc}
\usepackage[T1]{fontenc}
\usepackage{graphicx}
\usepackage{amsmath}
\usepackage{amsfonts}
\usepackage{amssymb}
\usepackage{booktabs}
\usepackage{multirow}
\usepackage{wrapfig}
\usepackage{xcolor}
\usepackage{svg}
\usepackage{nicefrac}
\usepackage{microtype}
\usepackage{url}
\usepackage{hyperref}   

\newif\ifdraft\drafttrue
\ifdraft
  \newcommand{\todo}[1]{\textcolor{red}{\textbf{[TODO: #1]}}}
\else
  \newcommand{\todo}[1]{}
\fi

\iclrfinalcopy

\title{Selective Latent Thinking:  Adaptive Compression of LLM Reasoning Chains}

\author{%
  \textbf{Hui Xie}$^{1,*}$, \hspace{0.5em} \textbf{Jie Liu}$^{1,*,\dagger}$, \hspace{0.5em} \textbf{Ziyue Qiao}$^2$, \hspace{0.5em} \textbf{Joaquin Vanschoren}$^1$ \\
  $^1$Eindhoven University of Technology, Netherlands \\
  $^2$School of Computing and Information Technology, Great Bay University, China \\
  \texttt{xiehui3019@gmail.com}, \texttt{\{j.liu9, j.vanschoren\}@tue.nl}, \texttt{zyqiao@gbu.edu.cn} \\
}

\begin{document}

\maketitle
{\let\thefootnote\relax\footnotetext{$^*$Equal contribution. \quad $^\dagger$Corresponding author.}}

\begin{abstract}

Explicit chain-of-thought (CoT) reasoning substantially improves the reasoning ability of large language models (LLMs), but incurs high inference cost due to lengthy autoregressive traces.
Existing latent reasoning methods offer a promising alternative, yet they often treat reasoning as uniformly compressible, causing precision-critical intermediate steps to be overly compressed and thereby degrading reasoning accuracy.
In this work, we propose \textbf{Selective Latent Thinking (SLT)}, a reasoning paradigm that selectively compresses redundant reasoning spans into latent representations while preserving precision-critical spans as explicit CoT within the same reasoning trajectory.
Our primary instantiation uses a lightweight decoder to anticipate a short upcoming reasoning span and applies confidence-based gating to determine the longest span that can be reliably compressed; the resulting selective compression policy is further optimized with trajectory-level reinforcement learning to balance answer correctness against reasoning cost.
Experiments across four mathematical reasoning benchmarks show that which spans are compressed is critical: with the same compressor, compressing arbitrary spans reduces the GSM-Hard accuracy of Qwen3-4B from 54.7\% to 17.3\%, whereas keeping numerical tokens explicit retains 51.8\%.
SLT reduces explicit reasoning length by 58.4\% with only a 2.8-point accuracy drop compared to explicit CoT on Llama-3.2-1B.
Notably, the principle extends beyond this design: instantiated within existing latent reasoning models via a learned compression gate, it improves accuracy at equal reasoning length (+2.6 points over Latent-SFT) and at shorter length (+5.6 points over Latent-GRPO with 13\% fewer tokens).
\end{abstract}
\section{Introduction}
As Large language models (LLMs) continue to scale, explicit chain-of-thought (CoT) reasoning~\citep{wei2022chain,xiang2025towards} has become a foundational paradigm for solving complex reasoning tasks. 
Recent advances further highlight its effectiveness when combined with reinforcement learning over long reasoning trajectories~\citep{jaech2024openai,guo2025deepseek,team2025kimi}, yielding substantial improvements in reasoning capability.
However, the inherent verbosity of CoT introduces high inference latency and substantial memory overhead, limiting efficiency and scalability~\citep{sui2025stop,feng2025efficient,wang2026render}.

Recent advances have sought to address this challenge by explicitly compressing CoT. One line of research operates at the token level, for example by identifying and skipping less informative tokens~\citep{xia2025tokenskip,zhang2025lightthinker,han2025token}, encouraging models to produce more concise reasoning traces~\citep{xu2025chain,aytes2025sketch,aggarwal2025l1,luo2025o1}, or terminating reasoning early once a sufficiently confident answer has been reached~\citep{yang2025dynamic}.
While effective, these methods remain confined to explicit token-level reasoning. A more promising alternative is to improve efficiency by performing reasoning process within dense latent spaces.
Pioneering works such as Coconut~\citep{hao2024training} and CODI~\citep{shen2025codi} laid the foundation for latent reasoning by distilling explicit CoT into continuous latent representations.
More recent methods, such as CoLaR~\citep{tan2025think} and RoT~\citep{wang2026render}, further extend this paradigm through dynamic latent compression and alternative latent reasoning forms.
However, these methods typically adopt a \textbf{homogeneous compressibility assumption}, where intermediate reasoning spans are regarded as uniformly suitable for latent compression.
This assumption does not always hold in precision-sensitive settings such as mathematical reasoning, where certain critical reasoning spans often need to remain explicit in order to preserve correctness.
\begin{figure}[!t]
    \centering
    \includegraphics[width=\textwidth]{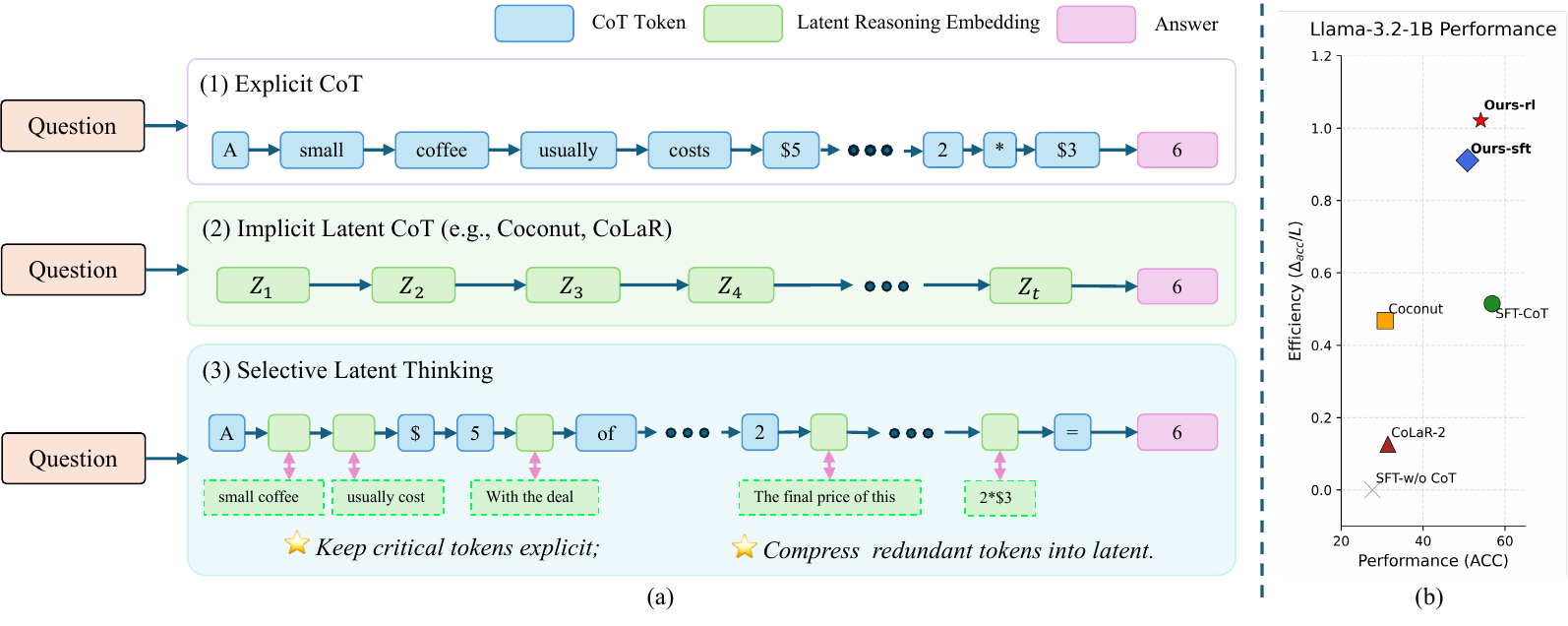}
    \vspace{-8mm}
    \caption{\textbf{Comparison of reasoning paradigms and efficiency analysis.} (a) Explicit CoT relies on verbose textual generation, whereas implicit latent CoT compresses the reasoning into latent space. Our \textbf{Selective Latent Thinking} (SLT) instead dynamically interleaves explicit CoT and latent reasoning, preserving precision-critical tokens in explicit form while selectively compressing redundant reasoning into latent. (b) SLT achieves a superior accuracy--efficiency trade-off ($\Delta_{acc}/L$) compared with both explicit CoT and implicit latent reasoning baselines.}
    \label{fig:motivation}
    \vspace{-5mm}
\end{figure}

To address this limitation, we introduce \textbf{Selective Latent Thinking (SLT)}, a reasoning paradigm that \emph{selectively compresses redundant reasoning spans into latent representations while preserving precision-critical spans as explicit CoT} within a single reasoning trajectory (Figure~\ref{fig:motivation}(a)).
In its primary instantiation, SLT uses a lightweight feature decoder to anticipate upcoming reasoning spans and a confidence gate to select the longest span that can be reliably compressed; the accepted span is encoded into a single latent representation, while uncertain spans fall back to explicit CoT.
This policy is learned in three stages: latent compression learning, reliability-aware feature decoding, and trajectory-level reinforcement learning that trades off answer correctness against explicit reasoning cost.
To examine whether the benefit stems from selectivity itself rather than from this particular design, we further derive \textbf{Latent-SLT}, which adds a learned compression gate to existing continuous-thought models (Section~\ref{sec:appendix_latent_slt}).

We evaluate SLT on four mathematical reasoning benchmarks under two rationale formats (GSM8k-AugNL and GSM8k-Aug), with Llama-3.2-1B and Qwen3-4B backbones.
On Llama-3.2-1B, SLT reduces explicit reasoning length by 58.4\% with only a 2.8-point accuracy drop relative to explicit CoT, comparing favorably with prior latent reasoning baselines (Figure~\ref{fig:motivation}(b)).
Latent-SLT further improves existing latent reasoning models at equal or shorter reasoning length (e.g., +5.6 points over Latent-GRPO with 13\% fewer tokens).
Our contributions are:

\begin{itemize}
    \item We identify the homogeneous compressibility assumption of existing latent reasoning methods as a key limitation, and propose \textbf{SLT}, a reasoning paradigm that decides span by span whether to reason latently or explicitly.
    \item We develop a framework that realizes SLT through reliability-aware feature decoding and confidence gating, further optimized with trajectory-level reinforcement learning.
    \item Experiments and ablations show that which spans are compressed is decisive for accuracy, that SLT achieves a favorable accuracy--length trade-off, and that the principle transfers to existing continuous-thought models.
\end{itemize}
\section{Related Work}
\label{sec:related}
\paragraph{Efficient Reasoning with Explicit CoT Reduction.}
A related line of work improves reasoning efficiency by reducing the cost of explicit Chain-of-Thought (CoT) generation. Some methods operate at the token level, for example by skipping less informative tokens or encouraging shorter reasoning traces~\citep{xia2025tokenskip,xu2025chain,aytes2025sketch,aggarwal2025l1,luo2025o1}. Others reduce unnecessary reasoning at the trajectory level through adaptive stopping or budget allocation~\citep{yang2025dynamic,sui2025stop,feng2025efficient}. While effective, these approaches remain within the explicit token space and primarily aim to shorten reasoning trajectories. In contrast, our method reduces reasoning cost by selectively replacing compressible reasoning spans with latent representations, while preserving explicit CoT for precision-critical computation.

\paragraph{Latent Reasoning and Continuous Thought.}
Complementary to explicit CoT reduction, another line of work improves reasoning efficiency by shifting the reasoning process into continuous latent representations. Early methods such as Coconut~\citep{hao2024training} and CoDi~\citep{shen2025codi} establish this paradigm by distilling explicit CoT segments into latent tokens. Subsequent work further develops this idea: Compressed CoT~\citep{shen2025efficient} represents implicit reasoning as dense latent sequences, while CoLaR~\citep{tan2025think} improves latent decodability through additional structural constraints. More recent approaches, including RoT~\citep{wang2026render} and OneLatent~\citep{lv2026onelatent}, broaden latent reasoning to multimodal settings by coupling LLM hidden states with visual features. Moreover, Latent-GRPO~\citep{deng2026latent} further extends reinforcement learning to continuous latent reasoning. Unlike these methods, which predominantly reason directly in latent space, our framework first anticipates a short explicit reasoning trajectory and then selectively compresses only the accepted span into a compact latent representation.

\paragraph{Multi-Token Prediction and Future Anticipation.}
Recent advances improve autoregressive decoding efficiency by predicting multiple future tokens or intermediate representations in parallel.
Representative examples include Medusa~\citep{cai2024medusa}, which augments the backbone model with multiple decoding heads, as well as multi-token prediction methods such as Better \& Faster Large Language Models via Multi-token Prediction~\citep{gloeckle2024better} and Your LLM Knows the Future~\citep{samragh2025your}, which leverage the model’s ability to anticipate several upcoming tokens from a shared prefix. At the feature level, EAGLE-3~\citep{li2025eagle} further demonstrates that predicting future representations can substantially reduce decoding latency. Collectively, these works show that future tokens and intermediate representations can be reliably anticipated from the current context. We build on this insight by using a lightweight feature decoder to roll out short candidate reasoning trajectories, which serve as the basis for deciding where latent compression is appropriate.



\section{Method}
\label{sec:method}

\subsection{Overview}
\label{subsec:overall_framework}
\begin{figure}[t]
    \centering
    \includegraphics[width=\linewidth]{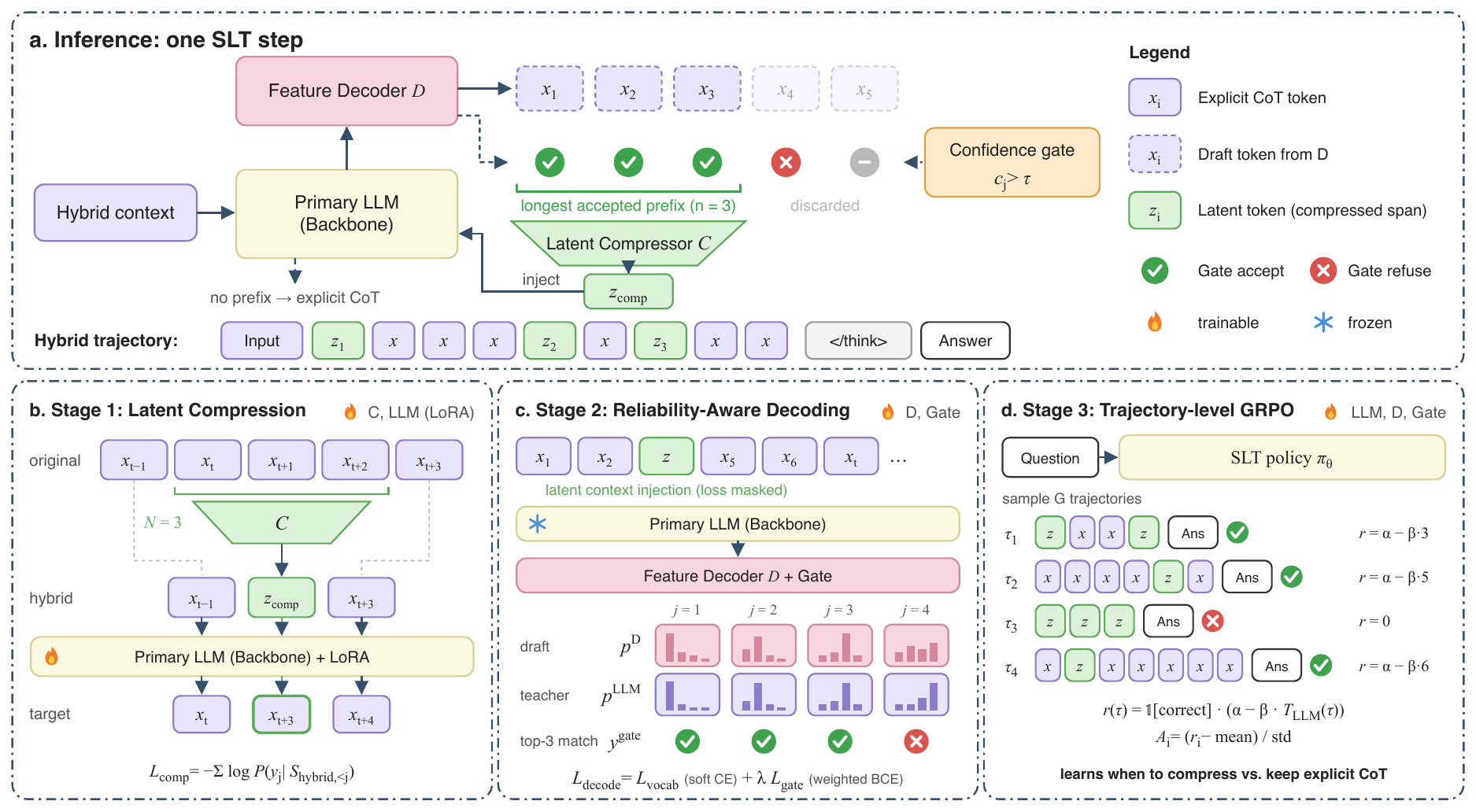}
    \vspace{-6mm}
    \caption{\textbf{Overview of the proposed Selective Latent Thinking (SLT).}
    (a) During inference, a lightweight decoder $\mathcal{D}$ drafts a short future reasoning trajectory, a confidence gate accepts its longest reliable prefix, and the latent compressor $\mathcal{C}$ maps the accepted span into a single latent token that is fed back to the LLM; when no prefix is accepted, SLT falls back to explicit CoT, yielding a hybrid trajectory of interleaved latent and explicit steps.
    (b) Stage~1 trains $\mathcal{C}$ (with LoRA on the LLM) to replace multi-step reasoning spans with single latent tokens under causal alignment.
    (c) Stage~2 jointly trains $\mathcal{D}$ and the gate on the frozen LLM to match its future token distributions and to predict which drafted steps are reliable enough for compression.
    (d) Stage~3 optimizes the full policy with trajectory-level GRPO, rewarding correct answers that use fewer explicit tokens.}
    \label{fig:model_arch}
    \vspace{-4mm}
\end{figure}

In this section, we introduce \textbf{Selective Latent Thinking (SLT)}, a framework that dynamically interleaves latent reasoning and explicit CoT within a single reasoning trajectory. The key idea is to replace only those intermediate reasoning spans that can be reliably compressed into latent representations, while preserving explicit reasoning form for the remaining steps. As shown in Figure~\ref{fig:model_arch}, SLT consists of three components: a primary large language model (LLM), a lightweight feature decoder $\mathcal{D}$, and a latent compressor $\mathcal{C}$.

During reasoning, the primary LLM maintains the current autoregressive state. Conditioned on the current hidden states of the primary LLM, the feature decoder $\mathcal{D}$ predicts a short future reasoning trajectory $x^{\mathcal{D}}_{1:J}$ that approximates the next few explicit CoT steps. A confidence gating module then evaluates this predicted trajectory and selects the longest continuous prefix that is reliable enough for latent compression. Formally, let $\mathbf{h}^{\mathcal{D}}_j$ denote the hidden state of $\mathcal{D}$ at step $j$. The gate projects this hidden state to a scalar confidence logit:
\begin{equation}
    c_j = \mathbf{w}_{\text{gate}}^\top \mathbf{h}^{\mathcal{D}}_j + b_{\text{gate}},
\end{equation}
and accepts the longest prefix of length $n \le K$ such that $c_i > \tau$ for all $i \le n$, where $\tau$ is a predefined threshold.
The accepted span is then encoded by the latent compressor $\mathcal{C}$ into a fixed-length latent:
\begin{equation}
    \mathbf{z}_{\mathrm{comp}} \in \mathbb{R}^{1 \times d}.
\end{equation}
This latent block is injected back into the primary LLM as a proxy embedding for the next autoregressive update, allowing multiple explicit reasoning steps to be replaced by a single latent transition. If no sufficiently reliable prefix is found, SLT simply continues with standard explicit CoT generation.

After \texttt{</think>}, both $\mathcal{D}$ and $\mathcal{C}$ are disabled, and the primary LLM generates the final answer through standard autoregressive decoding.

\subsection{Latent Compression Learning}
\label{subsec:stage1_compression}

We first train $\mathcal{C}$ to replace multi-step explicit reasoning spans with single latent transitions compatible with the primary LLM's autoregressive computation and causal structure.

We implement $\mathcal{C}$ as a cross-attention bottleneck. To preserve the sequential structure of the compressed span, we add learnable positional encodings to the token embeddings before compression. A learnable latent query then attends to these position-aware contextualized embeddings, producing a fixed-length latent representation. Given a reasoning span $\mathbf{x}_{t:t+N-1}$ of length $N$, the compressor outputs
\begin{equation}
    \mathbf{z}_{\mathrm{comp}} = \mathcal{C}(\mathbf{x}_{t:t+N-1}) \in \mathbb{R}^{1 \times d}.
\end{equation}

In practice, this bottleneck is effective at compressing smooth logical transitions, but less effective at preserving high-entropy discrete entities such as exact mathematical values. We empirically verify this limitation in Section~\ref{Effectiveness of the Latent Compressor}. Accordingly, this stage is designed to learn \emph{how} to compress reasoning spans, while the decision of \emph{when} latent compression should be applied is deferred to the reliability-aware mechanism introduced in the next stage.

\paragraph{Hybrid Sequence and Causal Alignment.}
To integrate latent representations into autoregressive decoding, we construct a hybrid sequence $\mathcal{S}_{\text{hybrid}}$ by replacing each contiguous reasoning span with a latent block of length $1$, whose embedding is given by $\mathbf{z}_{\mathrm{comp}}$. We then enforce causal alignment between the original sequence and the hybrid sequence. Specifically, if the compressed span begins at token $x_t$ and ends at $x_{t+N-1}$, the prefix state immediately before the latent block is trained to predict $x_t$, and the latent position is trained to predict $x_{t+N}$, i.e., the first token following the removed span.

We optimize the standard language modeling objective on the aligned hybrid sequence:
\begin{equation}
    \mathcal{L}_{\text{comp}} = - \sum_j \log P(y_j \mid \mathcal{S}_{\text{hybrid},<j}),
\end{equation}
where the targets $y_j$ follow the alignment rule above. This objective encourages the latent block to preserve the information necessary to bridge the compressed span and predict the continuation.

\subsection{Reliability-Aware Feature Decoding}
\label{subsec:stage2_decoding}

Latent compression learning equips the model with the ability to compress explicit reasoning spans into latent representations, but does not determine when latent compression is appropriate. Prior latent reasoning methods often rely on a \textbf{homogeneous compressibility assumption}, implicitly treating different parts of a reasoning trajectory as equally suitable for latent compression. In practice, however, the predictability of future reasoning varies substantially across spans, especially in precision-sensitive settings. To address this, we train a lightweight feature decoder with a confidence gate to estimate which predicted reasoning steps are reliable enough for latent compression.

Given the current hidden state of the primary LLM, the feature decoder $\mathcal{D}$ predicts a short future reasoning trajectory. Let $\mathbf{p}^{\mathcal{D}}_j$ denote the output distribution of $\mathcal{D}$ at step $j$, and let $\mathbf{p}^{\text{LLM}}_j$ denote the corresponding target distribution from the primary LLM. We train $\mathcal{D}$ by matching these distributions:
\begin{equation}
    \mathcal{L}_{\text{vocab}} = - \sum_j \sum_{v \in \mathcal{V}} p^{\text{LLM}}_{j,v} \log p^{\mathcal{D}}_{j,v}.
\end{equation}

To supervise the gate, we define a relaxed binary continuation label based on mutual top-$k$ inclusion:
\begin{equation}
    y^{\text{gate}}_j = \mathbb{I}\left[ \hat{x}^{\mathcal{D}}_j \in \mathcal{T}_3(\mathbf{p}^{\text{LLM}}_j) \land \hat{x}^{\text{LLM}}_j \in \mathcal{T}_3(\mathbf{p}^{\mathcal{D}}_j) \right],
\end{equation}
where $\hat{x}$ denotes the top-1 predicted token, and $\mathcal{T}_3(\mathbf{p})$ denotes the set of top-3 token indices from distribution $\mathbf{p}$. This relaxed criterion avoids penalizing predictions that differ at the top-1 token but remain highly consistent at the distribution level.

The gating module outputs a raw confidence logit $c_j$, which is optimized using a weighted binary cross-entropy with logits loss:
\begin{equation}
    \mathcal{L}_{\text{gate}} = - \sum_j w_j \left( y^{\text{gate}}_j \log \sigma(c_j) + (1-y^{\text{gate}}_j)\log(1-\sigma(c_j)) \right),
\end{equation}
where $\sigma$ is the sigmoid function used only for loss computation. The overall objective for this stage is
\begin{equation}
    \mathcal{L}_{\text{decode}} = \mathcal{L}_{\text{vocab}} + \lambda \mathcal{L}_{\text{gate}}.
\end{equation}

Joint training equips $\mathcal{D}$ to anticipate future reasoning and produce representations that support local compression reliability estimation.

\paragraph{Bridging the Train--Inference Gap via Latent Context Injection.}
During standard training, the feature decoder $\mathcal{D}$ is optimized purely on explicit text sequences, creating an exposure gap at inference time where the context consists of hybrid trajectories with interleaved latent states. To eliminate this discrepancy without introducing expensive offline verification, we adopt a latent context injection strategy in the final epoch of Stage~2 training. Specifically, a fixed proportion of intermediate reasoning spans in the context are directly replaced with their compressed continuous embeddings, while the cross-entropy loss over these latent token positions is masked out. This formulation ensures that both the feature decoder and the confidence gate learn to anticipate future steps conditioned naturally on preceding latent thoughts, bridging the training-inference gap while maintaining end-to-end efficiency.

\subsection{Trajectory-Level Accuracy--Cost Optimization}
\label{subsec:stage3_rl}

The first two stages establish local mechanisms for latent compression and reliability estimation, but they do not directly optimize the long-horizon consequences of compression decisions. A locally acceptable latent substitution may still harm the final answer if its errors propagate through the remaining reasoning trajectory. To address this long-horizon credit assignment problem, we formulate the full \textbf{Selective Latent Thinking (SLT)} framework as a sequential decision process and optimize the joint policy $\pi_\theta(\tau \mid x)$ using Group Relative Policy Optimization (GRPO).During RL, we introduce stochasticity into the gate, the feature decoder, and the LLM decoding, so that every decision carries a likelihood and policy gradients reach all modules.

We define a delayed reward that balances final-answer correctness against the cost of explicit token generation:
\begin{equation}
    r(\tau) = \mathbb{I}[\mathrm{Correct}(\tau)] \cdot \big( \alpha - \beta \cdot T_{\text{LLM}}(\tau) \big),
\end{equation}
where $\mathbb{I}[\mathrm{Correct}(\tau)] \in \{0,1\}$ indicates whether the final answer is correct, and $T_{\text{LLM}}(\tau)$ denotes the total number of CoT tokens generated by the primary LLM along trajectory $\tau$. In our implementation, we omit the conventional KL-divergence penalty against a reference model to reduce training overhead.
The optimization objective is:
\begin{equation}
    \mathcal{J}_{\text{RL}}(\theta) = \mathbb{E}_{\tau \sim \pi_\theta}[r(\tau)].
\end{equation}

Trajectory-level optimization allows the policy to account for the delayed effects of local decisions. Thus, the model learns when latent compression improves efficiency without harming correctness, and when explicit CoT should be preserved to maintain performance on precision-critical steps.

\section{Experiments}
\subsection{Experiment Settings}
\label{subsec:datasets}

\textbf{Datasets and Tasks.} We primarily train and evaluate our framework on \textbf{GSM8k-AugNL}, which contains roughly 385k training instances and over 1k test instances. We use this as our main setting because it better reflects the free-form natural-language CoT style commonly used in practice. To evaluate robustness, we additionally test on \textbf{GSM-Hard}, \textbf{SVAMP}, and \textbf{MultiArith}. We also report results on the original \textbf{GSM8k-Aug} dataset, whose more structured rationale format (e.g., tagged intermediate spans like ``<<5*2=10>><<30/10=3>>'') is better matched to prior latent reasoning baselines such as \textbf{CoLaR-2} and \textbf{Coconut}. To assess SLT's generality, we further apply selective thinking to \textbf{Latent-SFT} and \textbf{Latent-GRPO} under their original experimental settings.

\textbf{Baseline Methods.} We compare against several standard and state-of-the-art baselines. \textbf{SFT-w/o CoT} is trained by discarding reasoning steps from the dataset, forcing the model to output the final answer directly ($L=0$). \textbf{SFT-CoT} is the standard full-parameter fine-tuning baseline trained with complete CoT rationales. For efficient and latent reasoning baselines, we compare against \textbf{Coconut}, \textbf{CoLaR-2}, and \textbf{RoT}. We evaluate our framework in two settings: \textbf{Ours-SFT}, corresponding to the model after Sec.~\ref{subsec:stage2_decoding} training, and \textbf{Ours-RL}, corresponding to the final model after  reinforcement learning.
All reported lengths (\#L) represent reasoning CoT tokens only. As a coarse proxy for the accuracy--efficiency trade-off, we also report $\Delta_{acc}/L$, defined as the accuracy improvement over \textbf{SFT-w/o CoT} divided by CoT length.


\textbf{Implementation Details.} We employ Llama-3.2-1B and Qwen3-4B as our primary backbones. To establish strong reasoning baselines, both models are first fully fine-tuned on GSM8k-AugNL. For our framework, Stage 1 trains the latent compressor with LoRA using a progressive schedule that gradually increases the proportion of compressed CoT spans. Stage 2 freezes the primary LLM and trains the feature decoder and confidence gate. Stage 3 unfreezes all modules and performs joint optimization with reinforcement learning. For fair comparison, we applied greedy decoding  to all evaluated methods if applicable. Full hyperparameter settings are provided in the Appendix.

\begin{table*}[t]
\vspace{-4mm}
\centering
\caption{\textbf{Results across four math reasoning benchmarks}. Pass@1 is accuracy (\%), \#L is average reasoning length, and $\Delta_{acc}/L$ is the average accuracy gain over SFT-w/o CoT per reasoning token. $^\dagger$ marks results from the RoT paper. All Llama-3.2-1B results and Qwen3-4B results for SFT-w/o CoT and our models are evaluated locally.}
\resizebox{\textwidth}{!}{
\begin{tabular}{l|cc|cc|cc|cc|ccc}
\toprule
& \multicolumn{2}{c|}{GSM} & \multicolumn{2}{c|}{GSM-Hard} & \multicolumn{2}{c|}{SVAMP} & \multicolumn{2}{c|}{MultiArith} & \multicolumn{3}{c}{Average} \\
& Pass@1 & \# L & Pass@1 & \# L & Pass@1 & \# L & Pass@1 & \# L & Pass@1 & \# L & $\Delta_{acc}/L$ \\
\midrule
\rowcolor{blue!5} \multicolumn{12}{c}{\textit{\textbf{Llama-3.2-1B}}} \\
SFT-w/o CoT & 21.23 & 0.00 & 5.00 & 0.00 & 43.70 & 0.00 & 41.67 & 0.00 & 27.65 & 0.00 & - \\
SFT-CoT & 53.49 & 77.89 & 13.40 & 89.25 & 63.40 & 39.18 & 97.20 & 42.20 & 56.87 & 62.13 & 0.515 \\
Coconut & 24.24 & 6.00 & 5.49 & 6.00 & 40.70 & 6.00 & 52.55 & 6.00 & 30.74 & 6.00 & 0.468 \\
CoLaR-2 & 17.23 & 37.85 & 3.86 & 42.00 & 34.40 & 20.21 & 70.18 & 19.33 &  31.40 & 29.84 & 0.126 \\
\rowcolor[gray]{.9}\textbf{Ours-SFT} & 43.72 & 30.50 & 10.36 & 40.50 & 59.04 & 14.71 & 95.00 & 13.33 & 52.02 & 24.75 & 0.985 \\
\rowcolor[gray]{.9}\textbf{Ours-RL} & \textbf{46.47} & 32.09 & \textbf{11.52} & 43.35 & \textbf{61.60} & 13.88 & \textbf{96.67} & 14.18 & \textbf{54.07} & 25.87 & \textbf{1.021} \\
\midrule
\rowcolor{blue!5} \multicolumn{12}{c}{\textit{\textbf{Qwen3-4B}}} \\
SFT-w/o CoT & 38.44 & 0.00 & 14.18 & 0.00 & 72.90 & 0.00 & 93.33 & 0.00 & 54.71 & 0.00 & - \\
SFT-CoT & 82.18 & 85.20 & 50.23 & 117.30 & 84.20 & 36.80 & 100.00 & 40.90 & 79.15 & 70.05 & 0.348 \\
Coconut$^\dagger$ & 16.90 & 6.00 & 5.42 & 6.00 & 43.60 & 6.00 & 60.30 & 6.00 & 31.60 & 6.00 & -3.852 \\
CoLaR-2 & 50.67 & 42.73 & 12.71 & 50.40 & 54.72 & 28.44 & 84.90 & 26.55 & 50.70 & 37.03 & -0.200 \\
RoT$^\dagger$ & 37.80 & 32.00 & 14.10 & 32.00 & 72.70 & 32.00 & 97.20 & 32.00 & 55.40 & 32.00 & 0.022 \\
Tokenskip & 74.32 & 45.40 & 39.17 & 76.33 & 81.00 & 20.37 & 99.40 & 20.43 & 73.47 & 40.06 & 0.468 \\
\rowcolor[gray]{.9}\textbf{Ours-SFT} & 63.84 & 38.84 & 29.72 & 64.49 & 76.20 & 15.90 & 95.00 & 17.69 & 66.19 & 34.23 & 0.335 \\
\rowcolor[gray]{.9}\textbf{Ours-RL} & \textbf{75.28} & 38.36 & \textbf{35.10} & 63.90 & \textbf{81.10} & 14.75 & \textbf{99.44} & 15.70 & \textbf{72.73} & 33.17 & \textbf{0.543} \\
\rowcolor[gray]{.9}\textbf{Ours-RL-gate0.2} & \textbf{76.57} & 40.96 & \textbf{36.10} & 64.64 & \textbf{81.90} & 15.53 & \textbf{99.44} & 16.61 & \textbf{73.50} & 34.43 & \textbf{0.546} \\
\bottomrule
\end{tabular}
}
\label{tab:main_results}
\vspace{-4mm}
\end{table*}

\subsection{Main Results}
\label{subsec:main_results}

\paragraph{Comparison to baseline methods on GSM datasets.}
Table~\ref{tab:main_results} compares SLT with state-of-the-art baselines on four grade-school mathematical reasoning benchmarks. SLT consistently achieves a stronger accuracy--efficiency trade-off than prior latent reasoning methods. In particular, on Llama-3.2-1B, \textbf{Ours-RL} improves average accuracy by 23.04\% relative to CoLaR-2 while using fewer CoT tokens (25.87 vs.\ 29.84). Relative to explicit CoT, it reduces reasoning length by 58.4\% with only 2.8\% performance degradation. This advantage stems from SLT's ability to preserve explicit reasoning for precision-critical spans and compress redundant reasoning steps into latent representations, rather than compressing the reasoning trajectory uniformly into latent form.
On Qwen3-4B, \textbf{Ours-RL} improves average accuracy over RoT by 17.33\% at a comparable compression ratio, indicating that selective fallback to explicit CoT remains beneficial for precision-sensitive reasoning with stronger models. 
Furthermore, the consistent improvement of \textbf{Ours-RL} over \textbf{Ours-SFT}, e.g.,  54.07\% vs. 50.82\% on Llama-3.2-1B, demonstrates that trajectory-level reinforcement learning effectively turns locally trained compression and gating modules into a coherent selective compression policy.
Since the FLOP overhead of our auxiliary modules is negligible compared to the LLM backbone, this token reduction yields an actual wall-clock speedup (see Appendix~\ref{sec:appendix_inference_time}).
\begin{table}[t]
\centering
\small
\caption{\textbf{Evaluation of SLT with structured CoT supervision (top) and its extensions to Latent-SFT and Latent-GRPO (bottom)}. Pass@1 is accuracy (\%) and \#L is average reasoning length. $^\dagger$ marks results from the CoLaR paper; all others are evaluated locally.}
\vspace{0mm}
\resizebox{\columnwidth}{!}{
\begin{tabular}{l|cc|cc|cc|cc|cc}
\toprule
& \multicolumn{2}{c|}{GSM} & \multicolumn{2}{c|}{GSM-Hard} & \multicolumn{2}{c|}{SVAMP} & \multicolumn{2}{c|}{MultiArith} & \multicolumn{2}{c}{Average} \\
\textbf{Method} & Pass@1 & \#L & Pass@1 & \#L & Pass@1 & \#L & Pass@1 & \#L & Pass@1 & \#L \\
\midrule
SFT-CoT   & 52.77 & 25.20 & 13.04 & 34.60 & 64.90 & 11.30 & 97.22 & 14.30 & 56.98 & 21.35 \\
Coconut$^{\dagger}$   & 23.10 & 6.00  & 5.49  & 6.00  & 40.70 & 6.00  & 41.10 & 6.00  & 27.59 & 6.00 \\
CoLaR-2$^{\dagger}$  & 40.10 & 12.70 & 9.08  & 14.00 & 54.90 & 6.11  & 91.30 & 7.35  & 48.84 & 10.04 \\
\rowcolor[gray]{.9}\textbf{Ours-SFT}  & 42.46 & 13.42 & 9.33  & 22.35 & 58.30 & 6.43  & 95.00 & 5.73  & 51.27 & 11.98 \\
\rowcolor[gray]{.9}\textbf{Ours-RL}  & \textbf{47.23} & 13.23 & \textbf{10.77}  & 21.03 & \textbf{60.80} & \textbf{5.86}  & \textbf{98.89} & \textbf{5.68}  & \textbf{54.42} & 11.45 \\
\midrule
Latent-SFT & 47.55 & 14.90 & 10.25 & 16.50 & 53.22 & 8.20 & 90.80 & 9.30 & 50.46 & 12.22 \\
Latent-GRPO & 66.29 & 16.70 & 15.83 & 19.00 & 56.50 & 8.70 & 94.65 & 10.20 & 58.32 & 13.65 \\
\rowcolor[gray]{.9}\textbf{Latent-SFT-SLT} & 50.22 & 14.60 & 11.48 & 17.70 & 55.44 & 7.90 & 95.20 & 8.70 & 53.09 & 12.22 \\
\rowcolor[gray]{.9}\textbf{Latent-GRPO-SLT} & \textbf{77.07} & 15.50 & \textbf{17.97} & 15.40 & \textbf{61.54} & 8.00 & \textbf{99.07} & 8.60 & \textbf{63.91} & 11.88 \\
\rowcolor[gray]{.9}\quad\textit{acc-only (no constraint)} & 77.32 & 29.70 & 18.77 & 35.70 & 60.54 & 15.10 & 98.23 & 16.50 & 63.72 & 24.25 \\
\bottomrule
\end{tabular}
}
\label{tab:gsmaug_results}
\vspace{-2mm}
\end{table}

\paragraph{Results on the Original GSM8k-Aug Format.}
Table~\ref{tab:gsmaug_results} presents results on the original GSM8k-Aug rationale format, which contains more structured and equation-oriented reasoning than GSM8k-AugNL. This setting provides a controlled comparison with prior latent reasoning baselines such as Coconut and CoLaR-2, while evaluating whether SLT can compress mathematical reasoning beyond merely removing linguistic redundancy.
Even under this more structured format, SLT consistently outperforms latent reasoning baselines, surpassing CoLaR-2 by 5.58\% in average accuracy with comparable CoT length. Relative to explicit CoT, SLT incurs only a 2.56\% accuracy drop while reducing reasoning length by 46.27\%. This advantage is especially clear on MultiArith, where \textbf{Ours-RL} outperforms SFT-CoT (98.89\% vs.\ 97.22\%) while reducing CoT length from 14.30 to 5.68. These results highlight that SLT improves efficiency by selectively compressing mathematical reasoning itself, rather than simply reducing redundancy in free-form CoT.

\begin{table*}[t]
\vspace{-4mm}
\centering
\small
\caption{\textbf{Comparison of explicit and latent reasoning on Qwen2.5-Math-7B, including Latent-SFT-SLT}. We report Pass@1 accuracy (\%) and average reasoning length (\#L) across four benchmarks.}
\vspace{0mm}
\resizebox{\textwidth}{!}{
\begin{tabular}{l|cc|cc|cc|cc|cc}
\toprule
& \multicolumn{2}{c|}{Math500} & \multicolumn{2}{c|}{AIME24} & \multicolumn{2}{c|}{AIME25} & \multicolumn{2}{c|}{GPQA} & \multicolumn{2}{c}{Average} \\
\textbf{Method} & Pass@1 $\uparrow$ & \#L $\downarrow$ & Pass@1 $\uparrow$ & \#L $\downarrow$ & Pass@1 $\uparrow$ & \#L $\downarrow$ & Pass@1 $\uparrow$ & \#L $\downarrow$ & Pass@1 $\uparrow$ & \#L $\downarrow$ \\
\midrule
\multicolumn{11}{l}{\textit{Explicit Reasoning}} \\
SFT  & 67.40 & 2662 & 13.39 & 6423 & 6.77 & 6550 & 32.07 & 3555 & 29.91 & 4798 \\
GRPO & 75.15 & 2982 & 16.61 & 7500 & 19.79 & 6815 & 38.25 & 4568 & 37.45 & 5466 \\
\midrule
\multicolumn{11}{l}{\textit{Latent Reasoning (No Sampling)}} \\
Latent-SFT  & 64.85 & 1070 & 6.67 & 1651 & 6.61 & 1299 & 29.67 & 1214 & 26.95 & 1309 \\
Latent-GRPO & 80.40 & 929 & 26.56 & 2548 & 23.23 & 1735 & 36.68 & 1384 & 41.72 & 1649 \\
\rowcolor[gray]{.9}\textbf{Latent-SFT-SLT}  & 62.63 & 1035 & 7.00 & 1598 & 6.77 & 1335 & 28.43 & 1114 & 26.20 & 1264 \\
\rowcolor[gray]{.9}\textbf{Latent-GRPO-SLT} & 82.28 & 909 & 26.89 & 2279 & 25.81 & 1724 & 38.55 & 1272 & 43.38 & 1546 \\
\bottomrule
\end{tabular}
}
\label{tab:think_results}
\vspace{-2mm}
\end{table*}

\subsection{Generality of SLT}
\label{subsec:latent_slt}

To assess SLT's generality, we augment \textbf{Latent-SFT} and \textbf{Latent-GRPO} with a learned gate that adaptively selects between explicit reasoning and latent compression, deriving \textbf{Latent-SFT-SLT} and \textbf{Latent-GRPO-SLT}, respectively. We evaluate the resulting SLT variants against the original baselines on Llama-3.2-1B-Instruct under the Latent-GRPO protocol. Implementation details are provided in Appendix~\ref{sec:appendix_latent_slt}.
Table~\ref{tab:gsmaug_results} shows accuracy gains across all four benchmarks in both settings. \textbf{Latent-SFT-SLT} raises average accuracy from 50.46\% to 53.09\% at an unchanged average reasoning length of 12.22 tokens. \textbf{Latent-GRPO-SLT} improves average accuracy over Latent-GRPO by 5.59 percentage points (58.32\% to 63.91\%) while reducing reasoning length from 13.65 to 11.88 tokens. An accuracy-only variant achieves comparable accuracy (63.72\%) but uses roughly twice as many reasoning tokens (24.25), highlighting the importance of preserving compression during optimization. Together, these results support the broader applicability of selective thinking for improving the accuracy--efficiency trade-off of latent reasoning.
Table~\ref{tab:think_results} reports evaluations on Qwen2.5-Math-7B. 
In the SFT stage, \textbf{Latent-SFT-SLT} performs slightly worse, as partitioning the already limited training data for gate training restricts reasoning supervision. 
After RL, however, \textbf{Latent-GRPO-SLT} comprehensively outperforms vanilla Latent-GRPO across all benchmarks, achieving both superior accuracy and shorter reasoning length.

\begin{figure}[!t]
    \centering
    \includegraphics[width=0.88\textwidth]{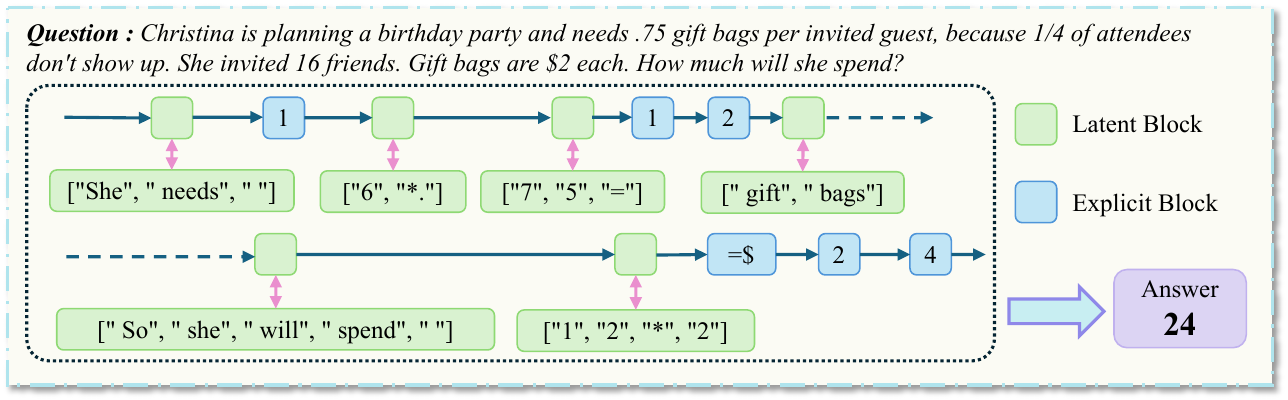} 
    \vspace{-3mm}
    \caption{\textbf{Example of SLT  reasoning trajectories.} SLT dynamically interleaves latent transitions with explicit reasoning while preserving precision-critical numerical and arithmetic steps.}
    \label{fig:sample_single_case}
    \vspace{-3mm}
\end{figure}


\subsection{Ablation Study \& Analysis}

\paragraph{Effects of Selective Compression.}
\label{Effectiveness of the Latent Compressor}
To test whether precision-critical spans should remain explicit, we compare SFT-CoT with two heuristics using the same trained compressor. \textit{Compress-Random} compresses randomly sized spans ($c \sim \mathcal{U}(2,5)$) into single latent embeddings, whereas \textit{Compress-SkipNum} preserves numeric tokens and compresses the surrounding text.
As shown in Table~\ref{tab:stage1_ablation}, \textit{Compress-Random} substantially reduces CoT length but severely degrades accuracy on precision-sensitive tasks, especially GSM-Hard. In contrast, \textit{Compress-SkipNum} recovers most of the lost performance while still maintaining a compact trajectory. This contrast shows that reasoning spans are not uniformly compressible, motivating SLT's reliability-aware gating mechanism to learn such compression decisions adaptively.
Examples in Figure~\ref{fig:sample_single_case} further demonstrate that SLT effectively compresses linguistic context while preserving precision-critical numerical and arithmetic steps.

\begin{figure}[!t]
    \centering
    \begin{minipage}{0.48\textwidth}
        \centering
        \includegraphics[width=\linewidth]{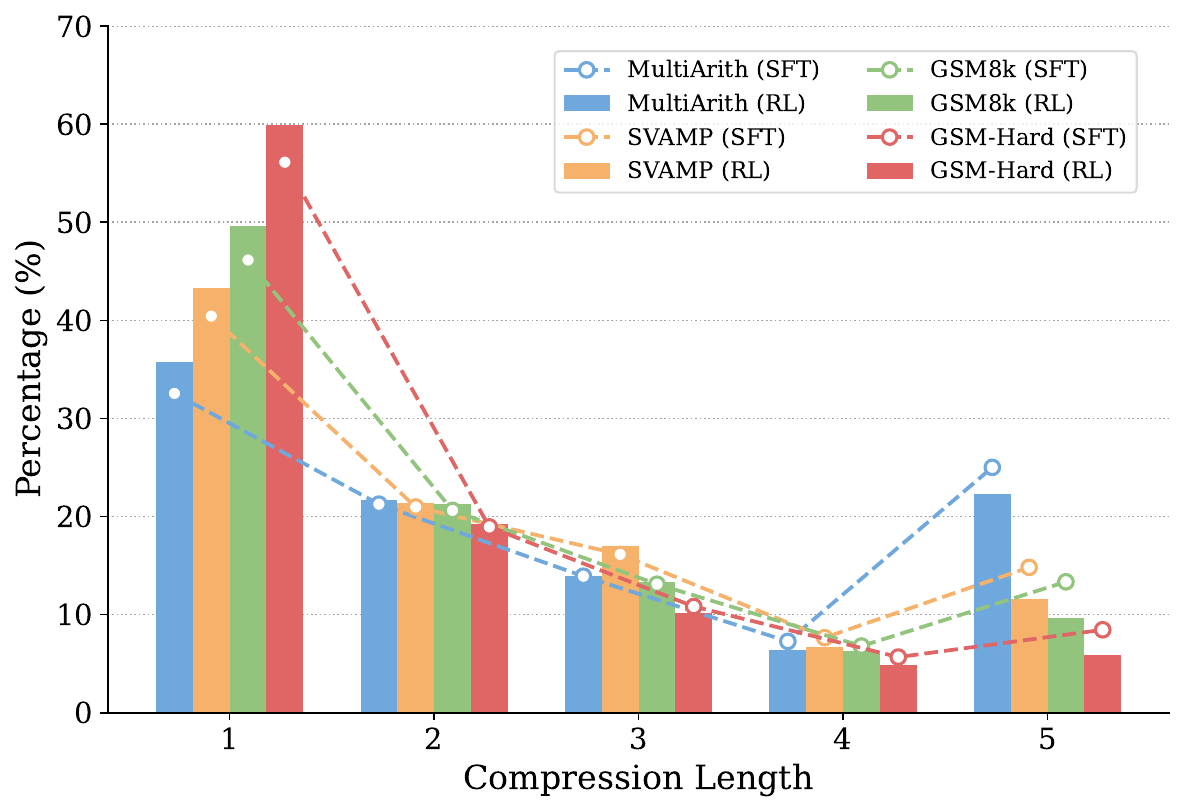}
        \vspace{-6mm}
        \caption{\textbf{Compression length distribution.} Compression lengths vary across datasets, with RL shifting the distribution toward shorter spans.}
        \label{fig:compression_llama}
    \end{minipage}
    \hfill 
    \begin{minipage}{0.46\textwidth}
        \centering
        \includegraphics[width=\linewidth]{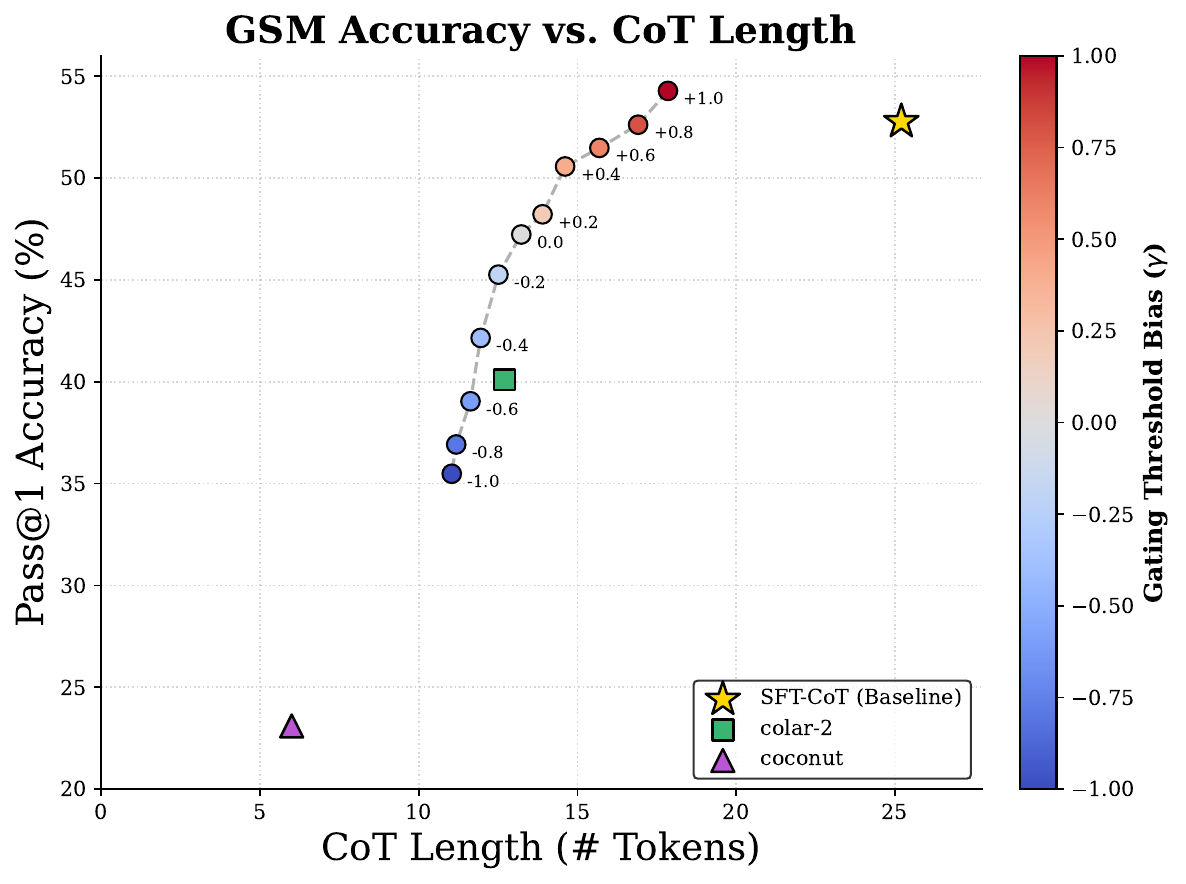}
        \vspace{-6mm}
        \caption{\textbf{Effect of gating threshold on GSM.} Adjusting the threshold controls the trade-off between accuracy and CoT length.}
        \label{fig:pareto_gsm}
    \end{minipage}
    \vspace{-4mm}
\end{figure}

\paragraph{Compression Policy Analysis.}
Figure~\ref{fig:compression_llama} visualizes the accepted compression lengths ($c=1$ to $5$). Overall, the policy generally favors shorter spans, gradually decreasing from $c=1$ to $4$, which reflects the frequent need to halt compression before unpredictable, precision-critical mathematical steps. The notable accumulation at $c=5$ is simply an artifact of capping safely compressible longer spans at the maximum look-ahead window. Comparing the policies, RL shifts the distribution towards a more conservative and robust regime: cautious short compressions ($c=1$) increase, while maximum-length compressions ($c=5$) decrease. Shorter local spans remain compatible with compact trajectories: RL improves accuracy while retaining comparable overall CoT lengths (Table~\ref{tab:main_results}). Avoiding unreliable long compressions may help limit state corruption and costly recovery.

\begin{table}[!t]
\vspace{-4mm}
\centering
\caption{\textbf{Stage 1 latent compressor evaluation.} ``Ratio'' denotes retained CoT length relative to SFT-CoT; $^\dagger$ marks ratios on GSM-Hard. \textit{Compress-Random} compresses randomly sized spans of 2--5 tokens, while \textit{Compress-SkipNum} preserves numeric tokens.}
\resizebox{\textwidth}{!}{
\begin{tabular}{l | cccc | cccc}
\toprule
\multirow{2}{*}{\textbf{Method}} & \multicolumn{4}{c|}{\textbf{Llama-3.2-1B}} & \multicolumn{4}{c}{\textbf{Qwen3-4B}} \\
\cmidrule(lr){2-5} \cmidrule(lr){6-9}
& \textbf{Ratio} & \textbf{SVAMP} & \textbf{MultiArith} & \textbf{GSM-Hard} & \textbf{Ratio} & \textbf{SVAMP} & \textbf{MultiArith} & \textbf{GSM-Hard} \\
\midrule
SFT-CoT & 100\% & 63.40 & 97.20 & 13.40 & 100\% & 85.50 & 100.00 & 54.70 \\
Compress-Random & $\sim$30\% & 60.50 & 98.30 & 10.90 & $\sim$30\% & 78.00 & 95.60 & 17.29 \\
Compress-SkipNum & $\sim$50\% (60\%$^\dagger$) & \textbf{63.40} & \textbf{97.78} & \textbf{13.30} & $\sim$57\% (80\%$^\dagger$) & \textbf{84.60} & \textbf{100.00} & \textbf{51.80} \\
\bottomrule
\end{tabular}
}
\label{tab:stage1_ablation}
\vspace{-2mm}
\end{table}

\begin{wraptable}{R}{0.55\textwidth}
\centering
\vspace{-3mm} 
\caption{\textbf{Ablation on Stage 2 joint gate training strategy}: (1) Ours-SFT: our complete joint training and refinement strategy; (2) w/o Joint Gate: isolated post-hoc refinement only; (3) w/o Gate: no confidence gating.}
\resizebox{\linewidth}{!}{
\begin{tabular}{lcccc}
\toprule
\textbf{Training Strategy} & \textbf{SVAMP} & \textbf{MultiArith} & \textbf{GSM} & \textbf{GSM-Hard} \\
\midrule
Ours-SFT & 76.20 & 95.00 & 63.84 & 29.72 \\
w/o Joint Gate & 72.40 & 93.89 & 57.77 & 27.90 \\
w/o Gate & 24.60 & 15.00 & 6.37 & 1.67 \\
\bottomrule
\end{tabular}}
\vspace{-4mm} 
\label{tab:gate_ablation}
\end{wraptable}

\paragraph{Effects of Confidence Gating Strategy.}
Table~\ref{tab:gate_ablation} evaluates confidence gating and its joint training with the feature decoder. Removing the gate entirely (\textit{w/o Gate}) leads to a severe performance drop, e.g., from 29.72 to 1.67 on GSM-Hard, showing that latent predictions cannot be accepted indiscriminately. Training the gate without joint optimization with the feature decoder (\textit{w/o Joint Gate}) also consistently degrades performance compared with \textit{Ours-SFT}. These results indicate that confidence gating is essential for reliable latent compression, and that joint training is necessary to align future trajectory prediction with compression reliability estimation.

\paragraph{Effects of Gating Threshold.}
Figure~\ref{fig:pareto_gsm} analyzes how the gating threshold modulates the trade-off between accuracy and CoT length. Lower thresholds accept longer predicted spans for latent compression, leading to shorter explicit reasoning trajectories, whereas higher thresholds impose stricter reliability requirements and preserve more explicit CoT.
Across thresholds, SLT traces a smooth Pareto frontier: accuracy improves steadily as more explicit reasoning is retained. The threshold thus provides a controllable operating point along the accuracy--efficiency trade-off. Compared with prior latent reasoning baselines, SLT achieves higher accuracy at comparable CoT lengths, and approaches the performance of full SFT-CoT while requiring substantially shorter reasoning trajectories.

\section{Conclusion}
\label{sec:concl}
We introduced \textbf{Selective Latent Thinking (SLT)}, a reasoning paradigm that challenges the homogeneous compressibility assumption of latent reasoning by interleaving latent and explicit CoT within a single reasoning trajectory.
Our primary instantiation anticipates upcoming reasoning spans, compresses only those deemed reliable by a confidence gate, and refines this policy with trajectory-level reinforcement learning.
Experiments show that which spans are compressed is decisive for accuracy, that SLT achieves a favorable accuracy--efficiency trade-off compared with prior latent reasoning baselines, and that the principle transfers to existing continuous-thought models via Latent-SLT.
These findings highlight the value of learning when to compress and when to reason explicitly.


\subsection*{AI use statement}
Generative AI tools were used for language polishing , LaTeX formatting,  research ideation, code development,  not for data generation. The authors reviewed the edits and take responsibility for the final manuscript.

\subsection*{Ethics statement}
This study uses existing reasoning benchmarks and pretrained models. SLT can still produce incorrect answers, and latent compression reduces the visibility of intermediate reasoning. Efficiency gains should not be interpreted as guarantees of correctness.

\subsection*{Reproducibility statement}
Datasets and evaluation settings are described in Section~\ref{subsec:datasets}. Training procedures and hyperparameters are detailed in Appendices~\ref{sec:appendix_latent_slt} and~\ref{subsec:training_setting}. Our code will be released publicly in the near future.

\bibliography{refs}
\bibliographystyle{iclr2027_conference}

\appendix

\newpage
\section{Latent-SLT}
\label{sec:appendix_latent_slt}

To demonstrate the generality of the selective thinking principle beyond the modular architecture in Section~\ref{sec:method}, we extend the framework to native continuous-thought models (specifically within the Latent-SFT and Latent-GRPO paradigm), denoted as \textbf{Latent-SLT}. This extension provides the \textbf{Latent-SFT-SLT} and \textbf{Latent-GRPO-SLT} variants evaluated in Table~\ref{tab:gsmaug_results}. In this setting, the model is trained via a two-stage pipeline: a supervised stage that aligns the dynamic gating mechanism with the continuous latent representations, followed by a reinforcement learning stage that optimizes the policy for end-to-end reasoning correctness. Specifically, \textbf{Latent-SFT-SLT} denotes the checkpoint obtained after the supervised stage, which also serves as the initialization for the subsequent RL training that yields \textbf{Latent-GRPO-SLT}. This section details the reinforcement learning formulation and training objectives of this extension.
\subsection{Supervised Stage}
\label{subsec:appendix_latent_slt_sft}

The supervised half produces the two objects the RL stage takes as given: the \emph{latent
vocabulary} --- the expert displacements $\mathbf{A}_k,\mathbf{B}_k$ of
Eq.~\eqref{eq:appendix_expert} and the ability to consume a top-$m$ mixture in place of a
token --- and the \emph{gate} $\mathrm{Gate}_\theta$ of Eq.~\eqref{eq:appendix_gate_mlp}.
They are produced in that order, by three separate procedures, because each depends on the
previous one being frozen.

\subsubsection{Latent Vocabulary}
\label{subsec:appendix_slt_stage1}

A chain of thought is first cut into consecutive segments whose lengths are the compression
values themselves. For a CoT of tokens $c_{1:N}$ we draw segment lengths
$k \in \{1,\dots,K\}$ i.i.d.\ from a fixed categorical distribution $\mathbf{w}$ and cut
accordingly, so a single training example contains segments of several different $k$ rather
than one global compression rate. Each segment is terminated by a reserved token
\texttt{<|compress\_token|>} on the encoder side; \emph{every} segment receives one,
including $k=1$. This is not cosmetic: with $k=1$ segments left unmarked the segment
boundaries are only implicit --- for $k \ge 2$ they are recoverable from the spacing of the
preceding compress tokens, and mixing in unmarked $k=1$ destroys that regularity. Omitting
the marker costs $58.5$ accuracy points on the stage-1 reconstruction task
($0.8552 \rightarrow 0.2699$, Table~\ref{tab:slt_sft_ablation}).

An encoder reads the marked sequence and its hidden state $\mathbf{h}_p$ at each compress
position is mapped back into the vocabulary simplex,
\begin{equation}
    \mathbf{q}_p = \mathrm{softmax}\bigl(\mathbf{h}_p \mathbf{W}_{\text{out}}^{\!\top}\bigr),
    \qquad
    \mathbf{z}_p \;=\; \sum_{i=1}^{m} q_{p,(i)}\, \mathbf{E}[v_{p,(i)}] ,
    \label{eq:appendix_slt_readout}
\end{equation}
where $(i)$ indexes the top-$m$ entries of $\mathbf{q}_p$ and $\mathbf{E}$ is the decoder
embedding matrix. Eq.~\eqref{eq:appendix_slt_readout} is Eq.~\eqref{eq:appendix_gumbel}
without the Gumbel perturbation: the supervised stage forms the mixture deterministically,
and RL later injects noise into the same construction. A latent block is therefore never an
unconstrained vector --- it is a convex combination of $m$ real token embeddings, which is
what keeps the latent space anchored to pretrained lexical semantics.

The decoder receives $\mathbf{z}_p$ in place of the segment it stands for and must continue
the reasoning from there. We use the \emph{continuation} layout: within a segment of length
$k$, one primary element decodes the following segment while the remaining $k-1$ elements
attend only to the segment anchor and to positions inside the segment; the last primary
element predicts \texttt{</think>}; the first segment is not decoded. Because a block
standing for two steps and one standing for three should not be the same distribution, each
$k \ge 2$ owns a low-rank expert $(\mathbf{A}_k,\mathbf{B}_k)$ applied to the output logits;
$k=1$ uses the bare head, since compressing a single token is the identity and needs no
mixing capacity.

Stage 1 is optimised in three consecutive runs rather than end to end --- the vocabulary-space
constraint makes the joint objective slow to converge --- and the best checkpoint of each
run initialises the next: \textbf{(i)} the encoder, \textbf{(ii)} the decoder conditioned on
the encoder's latent chains, \textbf{(iii)} joint refinement of both. Reconstruction accuracy
after the three runs is $0.9363$.

\subsubsection{Autonomous Latent Generation}
\label{subsec:appendix_slt_stage2}

The encoder is now discarded. The stage-1 system is run over the training set to emit, at
every latent position, the top-$m$ teacher distribution $(\mathbf{t}_p, \mathbf{v}_p)$; the
student is then trained to produce those distributions itself and to decode the answer,
\begin{equation}
    \mathcal{L} \;=\; w_{\text{ce}}\, \mathcal{L}_{\text{CE}}
                 \;+\; w_{\text{kl}}\, \mathcal{L}^{\text{top-}m}_{\text{KL}},
    \qquad
    \mathcal{L}^{\text{top-}m}_{\text{KL}}
    = \frac{1}{|P|}\sum_{p \in P} \sum_{i=1}^{m} t_{p,i}
      \bigl(\log t_{p,i} - \log \hat{p}_{p,v_{p,i}}\bigr),
    \label{eq:appendix_slt_stage2_loss}
\end{equation}
with $\mathcal{L}_{\text{CE}}$ the cross-entropy on the answer span, $P$ the latent
positions, and $\hat{\mathbf{p}}_p$ the student distribution normalised over the full
vocabulary. Ten independently diced segmentations of the same corpus are used as
augmentation, so the student sees the same CoT cut in different ways. The routed head
remains trainable here; the Gumbel perturbation of Eq.~\eqref{eq:appendix_gumbel} is already
active during this stage, so the student is trained under the same stochasticity it will
face at rollout time.

\subsubsection{Gate}
\label{subsec:appendix_slt_gate}

The gate cannot be trained jointly with Section~\ref{subsec:appendix_slt_stage2}, for a
reason that is worth stating because it dictates the data split. Its supervision must say
which $k$ each boundary can afford, and the natural signal is how much the ground-truth
continuation degrades when the boundary is compressed. On rows the student has been trained
on for tens of epochs that signal does not exist: the continuation is reproduced from
memory, and the measured degradation is numerically zero (median $0.001$ nats). We therefore
hold out the last $20{,}000$ rows of the corpus from Section~\ref{subsec:appendix_slt_stage2}
entirely --- the student never sees them --- and compute the gate's labels there.

On a held-out row, the trained student is frozen and each boundary $p$ is branched three
ways. For each $k$ the student's own output is used to continue --- the bare-head token for
$k=1$, the top-$m$ expert-$k$ mixture for $k \ge 2$ --- and we score the branch by the mean
cross-entropy of the ground-truth continuation that follows,
\begin{equation}
    k^{*}_p \;=\; \arg\min_{k \in \{1,\dots,K\}}
    \Bigl[\, \mathrm{score}_k(p) \;+\; \lambda\, \mathrm{cost}(k) \,\Bigr],
    \qquad
    \mathrm{cost}(k) = \bigl(1,\; 0.5,\; 0\bigr)_k ,
    \label{eq:appendix_slt_oracle}
\end{equation}
where $\mathrm{cost}$ charges for leaving reasoning explicit and $\lambda \ge 0$ is a single
compression knob. Because the three scores are stored per boundary, $\lambda$ is calibrated
afterwards on a grid without re-running the student; we pick the value giving a target mean
$k$ of $2.2$. This measurement is what makes the decision content-dependent, and it is the
step that matters: replacing $\mathrm{Gate}_\theta$ with the dice distribution $\mathbf{w}$
that generated the training segmentations --- same student, same token budget --- costs
$15.4$ points (Table~\ref{tab:slt_sft_ablation}).

The gate head of Eq.~\eqref{eq:appendix_gate_mlp} is then fit by cross-entropy on
$(\mathbf{h}_p, k^{*}_p)$ pairs, where $\mathbf{h}_p$ is the boundary hidden state the
student presents to its own output head. This costs about one GPU-hour and requires no
retraining of Section~\ref{subsec:appendix_slt_stage2}. On a $90/10$ split of the held-out
rows ($172{,}950$ train / $19{,}194$ validation boundaries) it reaches $0.675$ accuracy
against a $0.381$ majority-class baseline. The resulting checkpoint --- student, routed
experts and gate --- is \textbf{Latent-SFT-SLT}, and is the initialisation of
Section~\ref{subsec:appendix_latent_slt_rl}.

\begin{table}[h]
\centering
\small
\caption{Supervised stage. Both stage-1 runs and the stage-2 run share the LoRA and schedule
settings; they differ where marked. Reconstruction accuracy refers to the stage-1
encoder--decoder task, end-task accuracy to GSM8k-Aug.}
\begin{tabular}{lc}
\toprule
\textbf{Setting} & \textbf{Value} \\
\midrule
\multicolumn{2}{l}{\emph{Latent vocabulary (stage 1)}} \\
Segment lengths $k$ & $\{1,2,3\}$ \\
Segment length weights $\mathbf{w}$ & $1:2:2$ \\
Top-$m$ mixture width & 10 \\
Experts & $k \in \{2,3\}$; $k{=}1$ bare head \\
Layout & continuation, $k{=}1$ explicit on the primary track \\
Compress token after every segment & yes (incl.\ $k{=}1$) \\
Learning rate & $1\times 10^{-4}$ \\
Epochs (encoder / decoder / union) & 5 / 8 / 8 \\
Batch-size & 128 \\
\midrule
\multicolumn{2}{l}{\emph{Autonomous latent generation (stage 2)}} \\
Loss weights $w_{\text{ce}}$ / $w_{\text{kl}}$ & 1.0 / 1.0 \\
Diced segmentations per corpus & 10 \\
Routed head & trainable \\
Gumbel temperature / noise scale & 1.0 / 1.0 \\
Learning rate & $3\times 10^{-4}$ \\
Epochs & 60  \\
Batch-size & 128 \\
Gradient checkpointing & off \\
Held-out rows (never seen by the student) & 20{,}000 \\
\midrule
\multicolumn{2}{l}{\emph{Shared}} \\
LoRA rank / dropout & 64 / 0.1 \\
Warmup ratio / weight decay & 0.05 / 0.01 \\
LR schedule & cosine \\
Precision & bf16, ZeRO-1 \\
\midrule
\multicolumn{2}{l}{\emph{Gate}} \\
Hidden width & 512 \\
Cost vector $\mathrm{cost}(k)$ & $(1,\,0.5,\,0)$ \\
Target mean $k$ (calibrates $\lambda$) & 2.2 \\
Optimiser / lr / epochs / batch & AdamW / $1\times10^{-3}$ / 20 / 1024 \\
Train / validation boundaries & 172{,}950 / 19{,}194 \\
Gate accuracy (majority baseline) & 0.675 (0.381) \\
\bottomrule
\end{tabular}
\label{tab:slt_sft_hyperparameters}
\end{table}

\begin{table}[h]
\centering
\small
\caption{The two design decisions of the supervised stage, isolated. Upper block:
stage-1 reconstruction accuracy, showing that marking $k{=}1$ segments is what makes mixed
compression usable at all. Lower block: end-task accuracy and mean emitted length of the
same stage-2 student under three inference policies, showing that the gain comes from
\emph{learning} the compression decision rather than from compression itself.}
\begin{tabular}{lcc}
\toprule
\textbf{Configuration} & \textbf{Acc.} & \textbf{Len.} \\
\midrule
\multicolumn{3}{l}{\emph{Stage-1 reconstruction}} \\
$k \in \{2,3\}$, $1{:}1$ & 0.8560 & --- \\
$k \in \{1,2,3\}$, no compress token after $k{=}1$ & 0.2699 & --- \\
$k \in \{1,2,3\}$, compress token after every segment & 0.8552 & --- \\
\quad + three-run stage-1 schedule & \textbf{0.9363} & --- \\
\midrule
\multicolumn{3}{l}{\emph{Stage-2 student, inference policy}} \\
All explicit ($k{=}1$ everywhere) & 52.69 & 23.8 \\
Dice ($k \sim \mathbf{w}$, content-independent) & 35.18 & 14.3 \\
Gate (Eq.~\ref{eq:appendix_gate_mlp}) & \textbf{50.57} & \textbf{11.9} \\
\bottomrule
\end{tabular}
\label{tab:slt_sft_ablation}
\end{table}

\subsection{Reinforcement-Learning Stage}
\label{subsec:appendix_latent_slt_rl}

\subsubsection{The Gate as a Policy}
\label{subsec:appendix_gate_policy_sec}

At each reasoning step a gate head reads the hidden state $\mathbf{h}_t$ of the primary LLM
and emits a distribution over the number of steps to absorb,
\begin{equation}
    k_t \;\sim\; \mathrm{Gate}_\theta(\,\cdot \mid \mathbf{h}_t), \qquad k_t \in \{1,\dots,K\},
    \label{eq:appendix_gate_policy}
\end{equation}
where $k_t=1$ emits one ordinary token and $k_t \ge 2$ emits a single latent block standing
in for $k_t$ reasoning steps. A latent block advances the KV cache by one position, so
$k_t \ge 2$ is where the length saving comes from. The gate is a two-layer MLP with $K=3$:
\begin{equation}
    \mathbf{g}_t = \mathrm{GELU}\bigl(\mathbf{h}_t\mathbf{W}_1^{\!\top} + \mathbf{b}_1\bigr)\mathbf{W}_2^{\!\top} + \mathbf{b}_2 \in \mathbb{R}^{K},
    \qquad
    \log \mathrm{Gate}_\theta(k \mid \mathbf{h}_t) = \bigl[\log\mathrm{softmax}(\mathbf{g}_t)\bigr]_k .
    \label{eq:appendix_gate_mlp}
\end{equation}
Sampling $k_t$, rather than thresholding a confidence score, is what gives the compression
decision a likelihood and therefore a gradient.

\subsubsection{Where the Randomness Comes From}
\label{subsec:appendix_randomness}

Two further sources of randomness act inside a latent step. First, when $k_t \ge 2$ the
output logits are displaced by a frozen low-rank expert attached to that value of $k$,
\begin{equation}
    \boldsymbol{\ell}_t \;\leftarrow\; \boldsymbol{\ell}_t + s\,\bigl(\mathbf{h}_t\mathbf{A}_{k_t}^{\!\top}\bigr)\mathbf{B}_{k_t}^{\!\top},
    \label{eq:appendix_expert}
\end{equation}
so that a block standing for two steps and one standing for three are different
distributions rather than the same one reused. The experts $\mathbf{A}_k,\mathbf{B}_k$ and the
scale $s$ are not trained; gradient reaches them only through $\mathbf{h}_t$. Second, the
resulting distribution is restricted to its top-$m$ entries ($m=10$) and the mixture weights
are drawn with Gumbel-softmax,
\begin{equation}
    \mathbf{p}_t \;=\; \mathrm{softmax}\!\left(\frac{\log \mathbf{q}_t + \boldsymbol{\xi}_t}{T}\right),
    \quad \boldsymbol{\xi}_t \sim \mathrm{Gumbel}(0,1),
    \qquad
    \mathbf{z}_t \;=\; \sum_{i=1}^{m} p_{t,i}\, \mathbf{E}[v_{t,i}] ,
    \label{eq:appendix_gumbel}
\end{equation}
so what is fed back into the LLM is a convex combination of top-$m$ token embeddings rather
than a single embedding. Ordinary temperature sampling supplies the randomness at explicit
steps. The per-step log-likelihood therefore factorises into the compression decision and
the emission given that decision,
\begin{equation}
    \log \pi_\theta(a_t \mid x, \tau_{<t}) \;=\;
    \underbrace{\log \mathrm{Gate}_\theta(k_t \mid \mathbf{h}_t)}_{\text{compression decision}}
    \;+\;
    \underbrace{\log p_\theta^{(k_t)}(a_t \mid x, \tau_{<t})}_{\text{emission}} .
    \label{eq:appendix_logpi}
\end{equation}
The two terms are separable, which Section~\ref{subsec:appendix_rl_schedule} relies on.

\subsubsection{Objective}
\label{subsec:appendix_objective}

We optimise with GRPO. For each prompt $x$ a group of $G=8$ trajectories
$\{\tau_i\}_{i=1}^{G}$ is sampled from $\pi_{\theta_{\text{old}}}$, and the advantage is the
within-group standardisation of the reward,
\begin{equation}
    A_i \;=\; \frac{r(\tau_i) - \mathrm{mean}\bigl(\{r(\tau_j)\}_{j=1}^{G}\bigr)}
                   {\mathrm{std}\bigl(\{r(\tau_j)\}_{j=1}^{G}\bigr)} ,
    \label{eq:appendix_adv}
\end{equation}
which removes the need for a learned value network. Writing
$\rho_{i,t}(\theta) = \pi_\theta(a_{i,t}\mid x,\tau_{i,<t}) / \pi_{\theta_{\text{old}}}(a_{i,t}\mid x,\tau_{i,<t})$
for the per-step importance ratio and
\begin{equation}
    s_{i,t}(\theta) \;=\; \min\Bigl(\rho_{i,t}(\theta)\,A_i,\ \
    \mathrm{clip}\bigl(\rho_{i,t}(\theta),\,1-\epsilon,\,1+\epsilon\bigr)A_i\Bigr)
    \label{eq:appendix_surrogate}
\end{equation}
for the clipped surrogate, the objective is
\begin{equation}
    \mathcal{J}_{\text{RL}}(\theta) \;=\;
    \mathbb{E}\!\left[\frac{1}{G}\sum_{i=1}^{G}\frac{1}{|\tau_i|}\sum_{t=1}^{|\tau_i|}
    \begin{cases}
        s_{i,t}(\theta), & A_i \ge 0,\\[2pt]
        \max\bigl(s_{i,t}(\theta),\, c\,A_i\bigr), & A_i < 0,
    \end{cases}
    \right]
    \label{eq:appendix_grpo}
\end{equation}
with $\epsilon = 0.2$ applied symmetrically. The second branch is the dual-clip correction
of \citet{ye2020mastering}: when $A_i<0$ an importance ratio that has drifted far above one
would otherwise produce an unboundedly large negative term, and $c=3.0$ floors it. The
inner average is taken per trajectory and then across the group
(\texttt{seq-mean-token-mean}), so a long trajectory does not outweigh a short one. No KL
penalty against a reference policy and no entropy bonus are used. Groups in which every
trajectory receives the same raw accuracy are discarded before the update: their advantages
are identically zero by Eq.~\eqref{eq:appendix_adv}, so they contribute no gradient.

\subsubsection{The Two-Phase Schedule}
\label{subsec:appendix_rl_schedule}

Running Eq.~\eqref{eq:appendix_grpo} with the gate inside the policy is unstable in a
specific way, and the obvious remedy -- freezing the gate's weights -- does not work. The
gate's input is $\mathbf{h}_t$, a quantity the LLM is actively optimising. So long as
$\log\mathrm{Gate}_\theta(k_t\mid\mathbf{h}_t)$ appears in Eq.~\eqref{eq:appendix_logpi}, the
advantage reaches the compression decision through $\mathbf{h}_t$ whether or not
$\mathbf{W}_1,\mathbf{W}_2$ are pinned; pinning them only makes the path one step longer. And
the direction that path points is fixed, because an explicit step is never less expressive
than a latent block standing in for it: under a reward that tracks correctness alone, the
cheapest way to raise accuracy is to stop compressing. This, not the gate's parameters, is
what drives the latent share toward explicit CoT. We therefore split RL into two phases.

\paragraph{Phase 1: the gate acts, but is not credited.}
The gate still \emph{samples} $k_t$ at every reasoning step, so latent blocks are produced
exactly as at inference and the expert displacement of Eq.~\eqref{eq:appendix_expert} stays
active. What is removed is the first term of Eq.~\eqref{eq:appendix_logpi}:
\begin{equation}
    \log \pi_\theta(a_t \mid x, \tau_{<t}) \;=\; \log p_\theta^{(k_t)}(a_t \mid x, \tau_{<t})
    \qquad\text{(Phase 1)} ,
    \label{eq:appendix_logpi_phase1}
\end{equation}
i.e.\ the drawn $k_t$ is treated as exogenous -- part of the environment, in the same sense
that the sampling temperature is. The gate head additionally receives no gradient of its
own. Nothing else changes: the rollout still draws $k_t$ from the gate conditioned on
$\mathbf{h}_t$, and $k_t$ is still carried to the trainer alongside the sampled trajectory.
Because $\pi_{\theta_{\text{old}}}$ and $\pi_\theta$ are evaluated by the same function, the
importance ratio in Eq.~\eqref{eq:appendix_grpo} remains self-consistent and no other part
of the objective needs adjusting.

The consequence is precise: the policy is credited for \emph{what it writes into a latent
block} but never for \emph{having chosen to compress}. With the gradient path through
$\mathbf{h}_t$ severed, no directional pressure on the gate remains, and the only route the
advantage has left to raise accuracy is to write better latents and better explicit tokens.
That is the slow route -- improvement per step is markedly smaller than what a policy
allowed to shift its compression pattern can show -- but the gain is capability rather than
a change of representation. The reward is plain accuracy,
\begin{equation}
    r_i \;=\; \mathbb{I}[\mathrm{Correct}(\tau_i)] .
    \label{eq:appendix_r_phase1}
\end{equation}
No length term is required, because the one mechanism that could trade latent steps away
for accuracy is exactly the one receiving no gradient. Note that the latent share is
nevertheless not constant: the gate reads $\mathbf{h}_t$, and $\mathbf{h}_t$ moves as the LLM
updates, so the share drifts even with the gate's own weights pinned. Phase 1 runs for 2500
steps and its final checkpoint initialises Phase 2.

\paragraph{Phase 2: trainable gate under a latent-share constraint.}
The gate head now becomes a trainable module with its own AdamW optimiser and learning rate,
kept outside the FSDP shard and pushed to the rollout sampler after each update, and both
terms of Eq.~\eqref{eq:appendix_logpi} contribute. To stop the drift described above we do
not add a fixed length bonus -- which merely trades one arbitrary constant for another --
but constrain the latent share directly. Let
\begin{equation}
    \ell_i \;=\; \frac{n^{\text{latent}}_i}{n^{\text{think}}_i}
    \label{eq:appendix_latent_share}
\end{equation}
be the fraction of reasoning positions of trajectory $\tau_i$ that are latent. We require
$\mathbb{E}[\ell] \ge \rho$ and solve the constrained problem by dual ascent, giving
\begin{equation}
    r_i \;=\; \mathbb{I}[\mathrm{Correct}(\tau_i)] \;-\; \lambda\,\bigl(1 - \ell_i\bigr),
    \label{eq:appendix_r_phase2}
\end{equation}
where $1-\ell_i$ is the explicit share and $\lambda \ge 0$ is the multiplier, updated once
per batch of size $B$ from the constraint violation,
\begin{equation}
    \lambda \;\leftarrow\; \max\!\left(0,\; \lambda + \eta\Bigl(\rho - \tfrac{1}{B}\textstyle\sum_{i=1}^{B} \ell_i\Bigr)\right),
    \qquad \rho = 0.73,\ \ \eta = 0.01 .
    \label{eq:appendix_dual}
\end{equation}
The multiplier is part of the checkpoint state, so it survives preemption and restarts. The
behaviour is self-correcting: when the batch latent share falls below $\rho$ the penalty
grows until compression is restored, and when it rises above $\rho$ the penalty decays and
accuracy takes over. In our runs $\lambda$ settles near $0.9$ with the training share held
at $\approx 0.70$ and the validation share at $\approx 0.73$, i.e.\ the constraint binds
rather than collapsing.

\subsubsection{How the Compression Pattern Moves}
\label{subsec:appendix_k_evolution}

Table~\ref{tab:k_evolution} tracks the gate's own behaviour across the schedule. The
latent share $\ell$ is what Eq.~\eqref{eq:appendix_dual} constrains; the remaining columns
break the same reasoning positions down by the value of $k$ the gate chose, since
$k=1$ positions are the explicit ones, and $k=2$ and $k=3$ are latent blocks absorbing two
and three steps respectively.

\begin{table}[h]
\centering
\small
\caption{Validation-set behaviour of the gate at each point of the schedule. $\ell$ is the
latent share $n^{\text{latent}}/n^{\text{think}}$; the $k$ columns give the share of
reasoning positions at each branching value, so they sum to $100\%$ and the last two sum to
$\ell$. Accuracy is validation Pass@1.}
\begin{tabular}{lccccc}
\toprule
& \textbf{Acc} & $\boldsymbol{\ell}$ & $\boldsymbol{k=1}$ & $\boldsymbol{k=2}$ & $\boldsymbol{k=3}$ \\
\midrule
Before RL (Latent-SFT-SLT)            & 50.22 & 73.8\% & 26.2\% & 34.1\% & 39.7\% \\
After RL Phase 1 (gate not credited)  & --    & 74.7\% & 25.3\% & 30.2\% & 44.5\% \\
\textbf{After RL Phase 2}             & \textbf{77.07} & \textbf{66.8\%} & \textbf{33.2\%} & \textbf{57.8\%} & \phantom{0}\textbf{9.0\%} \\
\midrule
\textit{Control:} acc-only, gate trainable throughout & 77.32 & 16.9\% & 83.1\% & \phantom{0}8.6\% & \phantom{0}8.3\% \\
\bottomrule
\end{tabular}
\label{tab:k_evolution}
\end{table}

Two things are visible here that the latent share alone hides. First, selectivity survives
training: accuracy rises from $50.22$ to $77.07$ while $\ell$ stays high, $73.8\%$ before RL
and $66.8\%$ after Phase~2. Second, and less obviously, the \emph{composition} of that
share does not stay put. Phase~1 pushes $k=3$ up, from $39.7\%$ to $44.5\%$ -- with no gradient
on the gate, the drift simply follows $\mathbf{h}_t$. By the end of Phase~2 it has
collapsed to $9.0\%$, while $k=2$ has grown from $30.2\%$ to $57.8\%$. The latent share says
nothing about how deep each latent block is, and the policy buys accuracy by retreating to
shallower blocks rather than by giving up blocks. The control row makes the contrast sharp. It is a
separate run in which the gate is trainable from step~$0$ and the reward is accuracy alone,
and it matches the two-phase schedule on accuracy, $77.32$ against $77.07$. The two reach
that point differently: the two-phase policy keeps $66.8\%$ of its reasoning positions latent,
the control keeps $16.9\%$ and writes $83.1\%$ of them out explicitly -- the same accuracy at
roughly twice the reasoning length (Table~\ref{tab:gsmaug_results}). Accuracy alone does not
distinguish the two solutions, which is why the latent share has to be constrained rather than
left to the reward.

\begin{table}[h]
\centering
\small
\caption{Configuration for the latent RL runs. Both phases share every setting below; they
differ only in the gate and in the reward, Eq.~\eqref{eq:appendix_r_phase1} versus
Eq.~\eqref{eq:appendix_r_phase2}.}
\begin{tabular}{lc}
\toprule
\textbf{Setting} & \textbf{Value} \\
\midrule
Group size $G$ (rollouts per prompt) & 8 \\
Prompt batch size & 64 \\
Max response length & 128 tokens \\
Clip range $\epsilon$ (symmetric) & 0.2 \\
Dual-clip constant $c$ & 3.0 \\
Loss aggregation & \texttt{seq-mean-token-mean} \\
KL penalty to a reference policy & none \\
Entropy bonus & 0 \\
Gate branching factor $K$ & 3 \\
Top-$m$ mixture width & 10 \\
Group filter & raw accuracy \\
\midrule
Phase 1 length & 2500 steps \\
Phase 2 length & 5000 steps \\
Phase 2 target latent share $\rho$ & 0.73 \\
Phase 2 dual step size $\eta$ & 0.01 \\
Phase 2 initial multiplier $\lambda_0$ & 0 \\
\bottomrule
\end{tabular}
\label{tab:rl_hyperparameters}
\end{table}

\newpage

\section{Catastrophic Forgetting}
\label{sec:appendix_forgetting}

Compressing reasoning into latent representations changes what the model writes, and a
natural worry is that it also damages what the model knows. We therefore evaluate the
general-purpose ability of the trained models on three benchmarks that have nothing to do
with mathematical reasoning: MMLU, ARC-Challenge, and HellaSwag. Parenthesised numbers are
the change relative to the untrained backbone.

\begin{table}[h]
\centering
\small
\caption{General-ability retention after latent-reasoning training. Deltas are relative to
the untrained backbone. A model that had forgotten everything would sit at the random
baseline.}
\begin{tabular}{lccc}
\toprule
\textbf{Backbone} & \textbf{MMLU} & \textbf{ARC-C} & \textbf{HellaSwag} \\
\midrule
Llama-3.2-1B-Instruct (untrained) & 46.07 & 37.63 & 61.64 \\
CoLaR (official GSM checkpoint)   & 25.44 \small{($-20.63$)} & 32.34 \small{($-5.29$)} & 50.80 \small{($-10.84$)} \\
\rowcolor[gray]{.9}\textbf{Ours-RL} & \textbf{42.26} \small{($-3.81$)} & \textbf{38.48} \small{($+0.85$)} & \textbf{61.93} \small{($+0.29$)} \\
\midrule
Random baseline & 25.00 & 25.00 & 25.00 \\
\bottomrule
\end{tabular}
\label{tab:forgetting}
\end{table}

The contrast is stark. CoLaR, which compresses uniformly, has fallen to the random baseline
on MMLU --- $25.44$ against a chance level of $25.00$, i.e.\ its general knowledge is gone,
not merely degraded --- and has lost ground on the other two benchmarks as well. Our model
loses $3.81$ points on MMLU and is unchanged on ARC-Challenge and HellaSwag, both of which
land marginally above the untrained backbone.

We read this as a consequence of selectivity rather than a separate property. Uniform
compression forces every reasoning step through a continuous bottleneck, so the mapping from
hidden states to the vocabulary is rewritten everywhere, including on inputs that have
nothing to do with the training task. Selective compression leaves a substantial fraction of
positions decoding ordinary tokens, so that mapping is exercised throughout training and
survives it.

\newpage
\section{Inference Efficiency Analysis}
\label{sec:appendix_inference_time}

\begin{figure}[ht]
    \centering
    \includegraphics[width=0.9\linewidth]{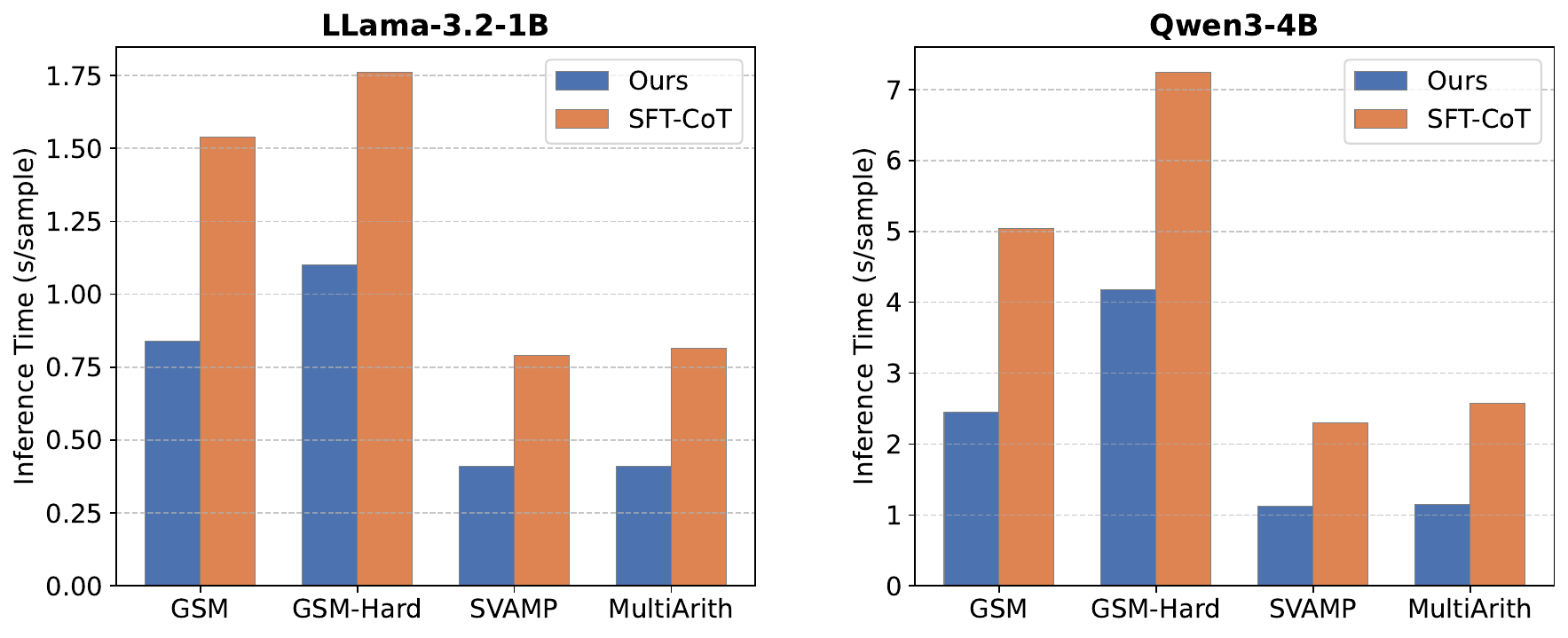}
    \caption{Average inference time comparison.Inference times for all models are measured on the GSM8k-AugNL dataset using a single A6000 GPU. SLT achieves substantial wall-clock speedups, particularly on the larger Qwen3-4B backbone, while dynamically adjusting computational effort on harder datasets like GSM-Hard.}
    \label{fig:inference_time}
\end{figure}

To evaluate practical efficiency, we compare the average wall-clock inference time of SLT against the explicit SFT-CoT baseline across four benchmarks (Figure~\ref{fig:inference_time}). SLT consistently accelerates inference on both backbones, but the speedup is notably more pronounced on Qwen3-4B. This discrepancy stems from the relative scale of the auxiliary modules. Our framework introduces a lightweight 1-layer feature decoder and a 2-layer compressor, which incur a similar, marginal absolute overhead on both models. However, because Qwen3-4B is significantly larger than Llama-3.2-1B, bypassing its heavy autoregressive generation yields much greater relative computational savings.

Furthermore, we observe that the relative speedup on GSM-Hard is smaller compared to SVAMP or MultiArith. This aligns perfectly with our observations in Section~\ref{subsec:main_results}: GSM-Hard contains extensive, precision-critical numerical reasoning where forced compression is unsafe. Consequently, our dynamic gating mechanism safely falls back to explicit CoT to preserve downstream accuracy. This deliberate trade-off naturally requires more explicit tokens, explaining the comparatively higher inference time on this specific dataset while avoiding the severe performance degradation seen in aggressive baseline methods.

\section{Inference Dynamics: Surpassing Explicit CoT via Threshold Modulation}
\label{sec:appendix_threshold_analysis}
\begin{figure}[htbp]
    \centering
    \includegraphics[width=\textwidth]{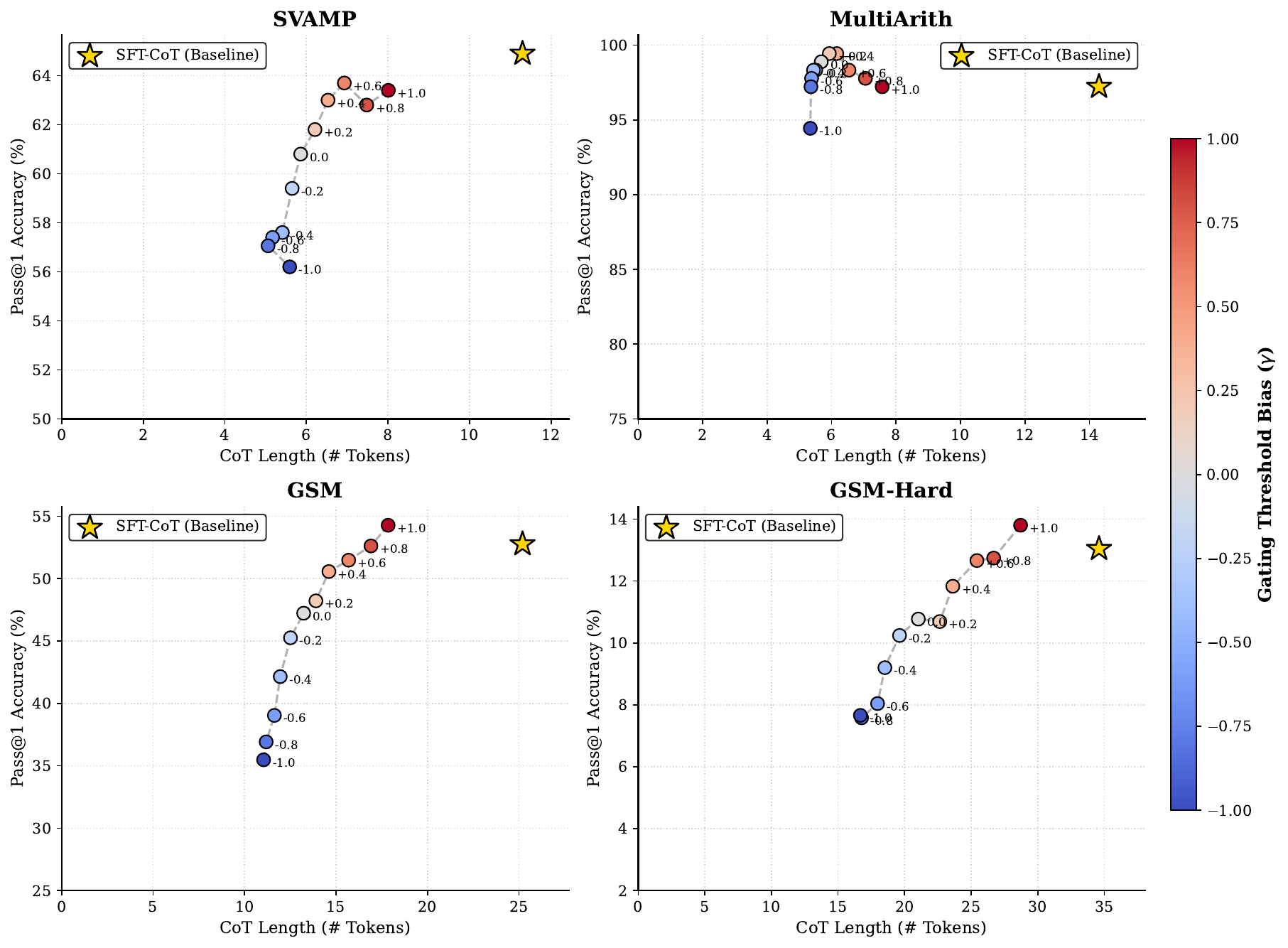} 
    \caption{Accuracy-efficiency Pareto frontier across different test datasets. By modulating the gating threshold bias $\gamma \in [-1.0, 1.0]$, SLT traces a smooth efficiency-accuracy trade-off. The gold star represents the SFT-CoT baseline. Notably, under conservative thresholds ($\gamma > 0$, red points), SLT frequently surpasses the explicit CoT accuracy ceiling while maintaining lower computational costs. Conversely, aggressive compression ($\gamma < 0$, blue points) exhibits diminishing marginal returns on length reduction and severe accuracy drops.}
    \label{fig:threshold_tradeoff}
\end{figure}
Complementing the main results in Section~\ref{subsec:main_results}, we explore whether inference-time threshold modulation can actively surpass the original SFT-CoT baseline. We trace the performance-efficiency Pareto frontier by varying the gating logit bias $\gamma \in [-1.0, 1.0]$. As illustrated in Figure~\ref{fig:threshold_tradeoff}, setting a conservative threshold ($\gamma > 0$) yields remarkable gains. Specifically, as $\gamma$ approaches $+1.0$, SLT actively outperforms the SFT-CoT baseline on critical benchmarks like GSM and GSM-Hard, while maintaining a substantially shorter token length. This indicates that selective compression acts as an implicit denoiser: by safely compressing linguistic fluff while explicitly preserving mathematical pivots, it mitigates the cascading hallucination errors typical of overly verbose explicit chains. 

Conversely, corroborating our theoretical intuition in Section~\ref{subsec:stage2_decoding}, pushing the framework into aggressive compression ($\gamma < 0$) reveals a stark cliff of diminishing marginal returns. As the threshold drops below zero, the horizontal gaps between points narrow significantly—indicating a physical limit to further length reduction—while accuracy drops precipitously. This confirms that forcing latent transitions beyond the model's intrinsic epistemic confidence aggressively destroys the underlying mathematical logic. Ultimately, this dynamic controllability allows practitioners to adopt a moderately conservative threshold ($\gamma \in [0.4, 1.0]$), unlocking an optimal operating regime that eclipses the explicit baseline's accuracy while consistently reducing computational cost.

\section{Implementation Details}
\label{subsec:training_setting}

To ensure reproducibility, the core architectural specifications and hyperparameter configurations across all training stages are summarized in Table~\ref{tab:hyperparameters}. We evaluate our framework on two prominent base models: \textbf{Llama-3.2-1B} and \textbf{Qwen3-4B}. 

\paragraph{Architecture \& Stage 1.} 
The latent compressor $\mathcal{C}$ is built with 2 cross-attention layers, while the feature decoder $\mathcal{D}$ uses a single transformer block. To build the representation foundation smoothly, Stage 1 employs a two-phase warmup strategy: we first freeze the primary LLM and train only the compressor for 3 epochs. Subsequently, we inject LoRA adapters ($r=128$, $\alpha=32$) into the LLM and jointly train for another 3 epochs.

\paragraph{Stage 2 \& Stage 3.} 
In the Stage 2 local dynamics alignment, we set the feature decoder's look-ahead window to $K=4$. Combined with the primary LLM's initial token, this creates a maximum latent span of 5 tokens, aligning perfectly with the capacity limits discussed in Appendix~\ref{sec:appendix_max_compression}. The loss components are balanced with a $1:2$ ratio, placing heavier emphasis on the gating module ($\mathcal{L}_{\text{gate}}$) to strictly penalize boundary errors. Finally, during the Stage 3 GRPO phase, we sample 8 distinct trajectories per prompt. The delayed reward strongly enforces exact factual correctness ($\alpha = 1.0$) while applying a conservative penalty for explicit token generation ($\beta = 0.01$ per token), ensuring the policy prioritizes logical rigor over reckless compression.

\begin{table}[htbp]
\centering
\small
\caption{Key hyperparameters and architectural configurations for the three-stage training pipeline. Note that the primary LLM is strictly frozen during Stage 2. For the Stage 3 RL optimization, Llama-3.2-1B undergoes full-parameter tuning, whereas Qwen3-4B is optimized via LoRA due to computational resource constraints.}
\begin{tabular}{llc}
\toprule
\textbf{Stage} & \textbf{Hyperparameter} & \textbf{Value} \\
\midrule
\multirow{4}{*}{\textbf{Stage 1}} 
& Phase 1 Training (Frozen LLM) & 3 Epochs \\
& Phase 2 Training (LoRA tuning) & 3 Epochs \\
& LoRA Configuration & $r=128$, $\alpha=32$ \\
& Learning Rate & $1 \times 10^{-4}$ \\
\midrule
\multirow{2}{*}{\textbf{Stage 2}} 
& Look-ahead Window ($K$) & 4 \\
& Training (Frozen LLM) & 3 Epochs \\
& Loss Weight ($\mathcal{L}_{\text{vocab}} : \mathcal{L}_{\text{gate}}$) & $1 : 2$ \\
& Learning Rate & $1 \times 10^{-4}$ \\
\midrule
\multirow{3}{*}{\textbf{Stage 3}} 
& GRPO Group Size & 8 \\
& Correctness Reward ($\alpha$) & 1.0 \\
& Token Penalty ($\beta$) & 0.01 \\
\bottomrule
\end{tabular}
\vspace{1mm}
\label{tab:hyperparameters}
\end{table}

\section{Interpretability and Visualization of Compression Trajectories}
\label{subsec:visualization_interpretability}

\begin{figure}[htbp]
    \centering
    \includegraphics[width=\textwidth]{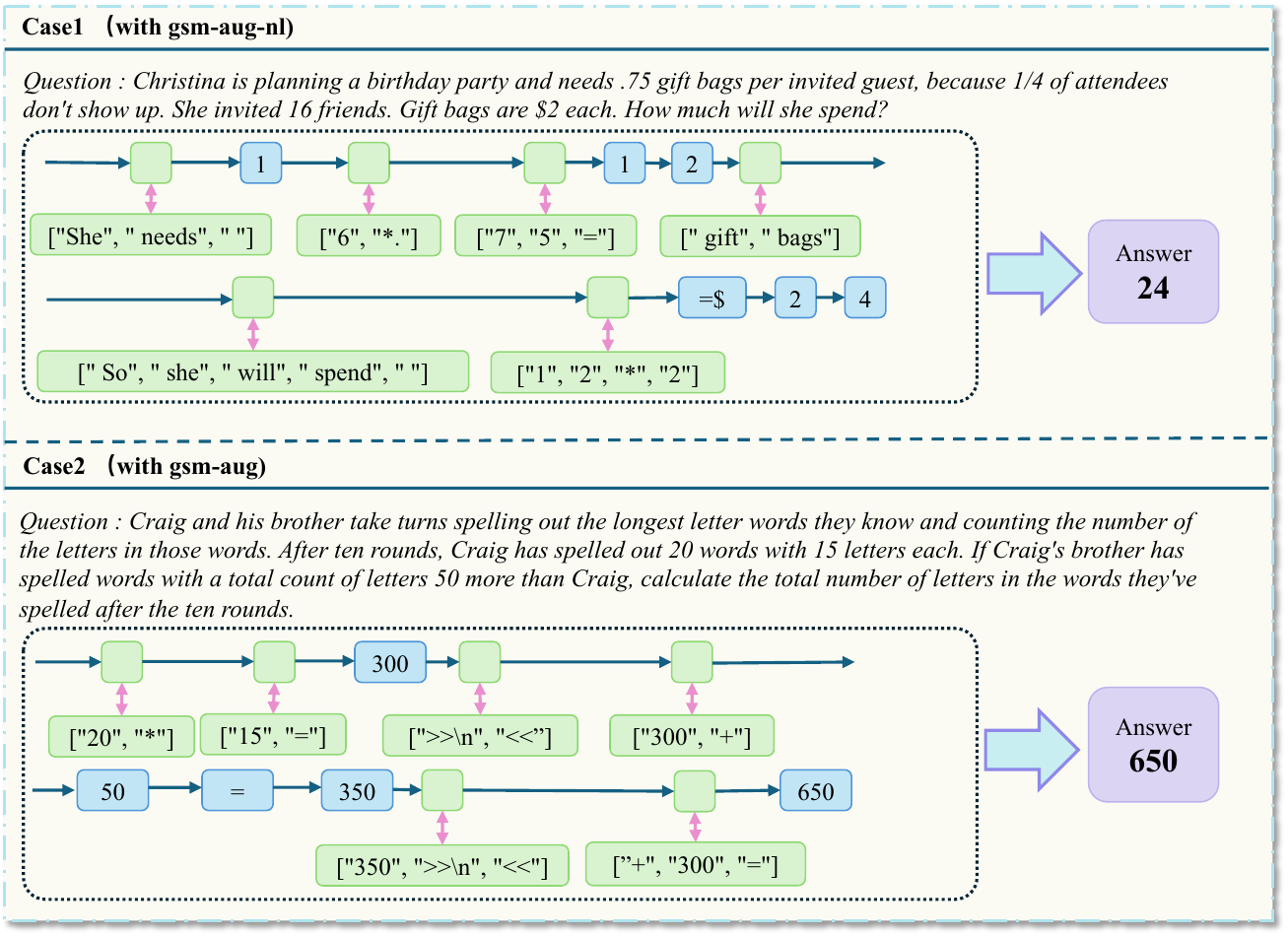} 
    \caption{Visualization of SLT generation trajectories. The timeline arrows illustrate the causal autoregressive flow. Green nodes represent compressed continuous latent vectors ($\mathbf{z}_{\mathrm{comp}}$) replacing the bracketed explicit strings below them. Blue nodes are explicit tokens triggered by the confidence gate's fallback mechanism, ensuring strict accuracy for numerical pivots.}
    \label{fig:sample_case}
\end{figure}

\paragraph{Visualization of Compression Trajectories.} 
To intuitively demonstrate the Selective Latent Thinking (SLT) mechanism, Figure~\ref{fig:sample_case} visualizes the actual generation trajectories. The directed arrows represent the autoregressive forward pass. Green nodes represent instances where multiple explicit tokens are functionally compressed into a single continuous latent vector ($\mathbf{z}_{\mathrm{comp}}$), while blue nodes indicate explicit tokens preserved by the confidence gate.

Across both natural language (Case 1) and structured equation (Case 2) formats, the model exhibits a clear "semantic routing" behavior. Predictable linguistic connective tissues (e.g., \texttt{["She", " needs", " "]}) and deterministic operational transitions (e.g., \texttt{["20", "*"]}) are consistently absorbed into compact latent blocks. Conversely, immediately prior to generating high-entropy mathematical pivots---such as exact intermediate values (\texttt{1}, \texttt{300}) or final answers (\texttt{24}, \texttt{650})---the gate aggressively halts compression. This forces the LLM to articulate fragile numerical entities explicitly. Crucially, the continuous autoregressive flow (blue arrows) confirms this is not extractive token skipping: the latent green nodes successfully carry the full semantic and causal payload required to accurately condition the subsequent explicit blue tokens.

\section{Ablation on Maximum Compression Boundary}
\label{sec:appendix_max_compression}

To unpack the truncation effect observed in Section~\ref{subsec:main_results}, we ablate the maximum look-ahead boundary $c_{\max}$ on the GSM benchmark using the Ours-SFT setting (Table~\ref{tab:cmax_ablation}). The results reveal a strict capacity limit for single-vector continuous representations. Pushing the boundary to $c_{\max} \ge 6$ causes a precipitous accuracy drop (down to 33.53\%). This highlights a severe \textbf{representation bottleneck}: forcing a single latent vector to simulate overly long discrete trajectories inevitably induces compounding uncertainty and destroys mathematical fidelity. Conversely, over-constraining the window ($c_{\max} = 3$) artificially fragments natural linguistic phrases, degrading both efficiency and accuracy due to suboptimal semantic chunking. Ultimately, the optimal manifold lies at $c_{\max} \in [4, 5]$ (peaking at 45.03\%), which perfectly scales to absorb predictable linguistic transitions while safely halting before critical high-entropy numerical pivots.

\begin{table}[h]
\centering
\small
\caption{Impact of varying the maximum compression boundary ($c_{\max}$) on GSM performance. Pushing $c_{\max} \ge 6$ degrades reasoning accuracy, revealing the intrinsic capacity limit of continuous representations.}
\begin{tabular}{ccc}
\toprule
\textbf{Maximum Boundary ($c_{\max}$)} & \textbf{Pass@1 Accuracy (\%)} & \textbf{Explicit CoT Length (\#L)} \\
\midrule
7 & 33.53 & 12.50 \\
6 & 36.09 & 12.43 \\
5 (Default) & 42.46 & 13.42 \\
4 & \textbf{45.03} & 14.78 \\
3 & 44.20 & 15.59 \\
\bottomrule
\end{tabular}
\vspace{1mm}
\label{tab:cmax_ablation}
\end{table}

\section{Why a Length Penalty on Explicit CoT Is Not Enough}
\label{sec:appendix_length_penalty}
A natural alternative to latent compression is to keep reasoning fully explicit and
simply penalise its length during RL. Table~\ref{tab:discrete-cot-tradeoff} reports
this baseline: GRPO on Llama-3.2-1B with an explicit length penalty in the reward.
Shortening does occur, but it is bought almost entirely with accuracy -- by step 50
the mean response length on GSM8K has dropped from 78.4 to 29.4 tokens while accuracy
falls from 53.9\% to 35.6\%. The degradation is uniform across benchmarks, including
the two that overlap the SFT training distribution.

\begin{table}[h]
\centering
\caption{Discrete CoT length-penalty GRPO on Llama-3.2-1B: accuracy (\%) and
mean response length (tokens). Reward $=\mathbb{1}[\text{correct}]-\lambda\,|\text{completion}|$,
$\lambda=0.005$, KL $\beta=0.04$; @$k$ denotes the GRPO checkpoint at step $k$.
$^{\dagger}$MultiArith and SVAMP overlap the (GSM8K-aug-NL) SFT training distribution;
GSM-Hard is out-of-distribution.}
\label{tab:discrete-cot-tradeoff}
\begin{tabular}{lcccccccc}
\toprule
& \multicolumn{2}{c}{GSM8K} & \multicolumn{2}{c}{GSM-Hard} & \multicolumn{2}{c}{MultiArith$^{\dagger}$} & \multicolumn{2}{c}{SVAMP$^{\dagger}$} \\
\cmidrule(lr){2-3}\cmidrule(lr){4-5}\cmidrule(lr){6-7}\cmidrule(lr){8-9}
Method & Acc & Len & Acc & Len & Acc & Len & Acc & Len \\
\midrule
SFT-CoT (base)   & 53.9 & 78.4 & 15.3 & 97.3 & 97.8 & 41.3 & 64.3 & 40.1 \\
\quad + GRPO @30 & 41.6 & 48.2 & 12.2 & 59.7 & 92.2 & 27.7 & 60.7 & 25.7 \\
\quad + GRPO @50 & 35.6 & 29.4 & \phantom{0}8.1 & 39.5 & 83.9 & 17.7 & 54.3 & 16.6 \\
\bottomrule
\end{tabular}
\end{table}
\section{Does Reinforcement Learning Improve CoLaR?}
\label{sec:appendix_colar_rl}

CoLaR is the another latent reasoning baseline in our comparison that supports reinforcement
learning, which raises the question of whether our advantage over it merely reflects the
fact that \textbf{Ours-RL} is RL-trained. It does not, for two reasons.

First, CoLaR's own paper reports no RL results on GSM. The reported CoLaR-2 number of
40.10 in Table~\ref{tab:gsmaug_results} is obtained with an MSE regression objective,
which makes latent generation deterministic and therefore leaves GRPO without the
sampling stochasticity it requires. Applying RL to CoLaR is only possible from its NLL
variant, which is already substantially weaker than the MSE variant before any RL is
applied (0.32 in the original paper; 0.26 in our reproduction, while our MSE reproduction
matches the published number).

Second, running RL from that NLL checkpoint with the official CoLaR code does not help.
Table~\ref{tab:colar_rl} reports the outcome: the per-benchmark changes are $+0.04$,
$-0.20$, $-1.00$ and $+0.36$, an average of $-0.20$. The comparison between CoLaR and our
method is therefore not a comparison of RL against SFT. Consistently with this, our
RL-free \textbf{Ours-SFT} already exceeds CoLaR-2 (MSE) on every benchmark.

\begin{table*}[t]
\centering
\small
\caption{Effect of reinforcement learning on CoLaR, under the original GSM8k-Aug rationale
format. ``CoLaR-2 (MSE, reported)'' is the published number, obtained with a deterministic
MSE objective that precludes GRPO; the NLL rows are the stochastic variant that RL requires,
trained and RL-tuned with the official CoLaR code. Each benchmark reports Pass@1 and CoT
length (\#L).}
\vspace{0mm}
\resizebox{\textwidth}{!}{
\begin{tabular}{l|cc|cc|cc|cc|cc}
\toprule
& \multicolumn{2}{c|}{GSM} & \multicolumn{2}{c|}{GSM-Hard} & \multicolumn{2}{c|}{SVAMP} & \multicolumn{2}{c|}{MultiArith} & \multicolumn{2}{c}{Average} \\
\textbf{Method} & Pass@1 & \#L & Pass@1 & \#L & Pass@1 & \#L & Pass@1 & \#L & Pass@1 & \#L \\
\midrule
CoLaR-2 (MSE, reported) & 40.10 & 12.70 & 9.08 & 14.00 & 54.90 & 6.11 & 91.30 & 7.35 & 48.84 & 10.04 \\
CoLaR-2 (NLL), SFT      & 23.31 & 13.40 & 5.34 & 43.60 & 43.16 & 9.40 & 82.77 & 7.30 & 38.65 & 18.43 \\
CoLaR-2 (NLL), + RL     & 23.35 & 13.70 & 5.14 & 44.70 & 42.16 & 9.50 & 83.13 & 7.40 & 38.45 & 19.33 \\
\midrule
\rowcolor[gray]{.9}\textbf{Ours-SFT} & 42.46 & 13.42 & 9.33 & 22.35 & 58.30 & 6.43 & 95.00 & 5.73 & 51.27 & 11.98 \\
\rowcolor[gray]{.9}\textbf{Ours-RL}  & \textbf{47.23} & 13.23 & \textbf{10.77} & 21.03 & \textbf{60.80} & \textbf{5.86} & \textbf{98.89} & \textbf{5.68} & \textbf{54.42} & 11.45 \\
\bottomrule
\end{tabular}
}
\label{tab:colar_rl}
\vspace{-2mm}
\end{table*}

\section{General Reasoning Results Beyond Mathematics}
\label{sec:appendix_2wiki}
\begin{wraptable}{r}{0.46\textwidth}
\vspace{-12pt} 
\centering
\small
\caption{Results on 2WikiMultiHopQA with Llama-3.2-1B. \#L denotes explicit CoT tokens only.}
\label{tab:2wiki}
\vspace{2pt}
\begin{tabular}{l|cc}
\toprule
\textbf{Method} & Pass@1 $\uparrow$ & \#L $\downarrow$ \\
\midrule
SFT-CoT & 34.8 & 48.00 \\
\rowcolor[gray]{.9}\textbf{Ours-SFT} & 33.8 & 19.95 \\
\rowcolor[gray]{.9}\textbf{Ours-RL} & \textbf{34.8} & \textbf{18.27} \\
\bottomrule
\end{tabular}
\vspace{-10pt} 
\end{wraptable}

All experiments in the main text are mathematical, which leaves open whether selective
latent thinking exploits redundancy specific to arithmetic rationales. To test this, we
evaluate SLT on \textbf{2WikiMultiHopQA}, a multi-hop question answering benchmark whose
reasoning chains are compositional rather than numeric, using the same Llama-3.2-1B
backbone and the same training pipeline as in Section~\ref{subsec:datasets}.

Table~\ref{tab:2wiki} reports the result. \textbf{Ours-RL} matches the explicit-CoT
baseline exactly (34.8 for both) while reducing CoT length from 48.00 to 18.27 tokens, a
reduction of 61.9\%. \textbf{Ours-SFT} is 1.0 point below the baseline at a 58.4\%
reduction. Selective compression is therefore not specific to mathematical reasoning: the
gate is able to identify which spans of a multi-hop derivation carry information that must
remain explicit, and which can be absorbed into latent blocks.

\section{Additional Results on Qwen2.5-7B}
\label{sec:appendix_qwen25_7b}

To examine whether our approach generalizes to other backbones, we further evaluate it on Qwen2.5-7B. In this setting, we collect the reasoning trajectories produced by TokenSkip on its training set and use them as training data to further train the model, following the same SFT and RL pipeline as in the main experiments. The results are reported in Table~\ref{tab:qwen25_7b_results}.

As shown in Table~\ref{tab:qwen25_7b_results}, SLT still achieves a substantial reduction in reasoning length at a relatively low cost in accuracy. Compared with Base CoT, Ours-RL reduces the average number of generated tokens from 280 to 80.28 (a 71.3\% reduction) while retaining 89.1\% of the average accuracy (75.62 vs.\ 84.85). Notably, although our model is trained on trajectories that have already been compressed by TokenSkip, it shortens them by a further 51.0\% (164 $\rightarrow$ 80.28), with an accuracy gap of only 4.45 points. On simpler benchmarks the degradation is marginal: on MultiArith, Ours-RL reaches 99.20\% with 45 tokens, compared with 100\% and 195 tokens for Base CoT; on SVAMP, it attains 87.60\% while using roughly one quarter of the tokens. The accuracy drop is more pronounced on GSM-Hard, suggesting that problems requiring longer computation chains remain more sensitive to aggressive compression. In addition, the RL stage improves over SFT by 4.78 points in average accuracy (70.84 $\rightarrow$ 75.62) while slightly reducing the reasoning length (81.58 $\rightarrow$ 80.28), indicating that RL recovers part of the accuracy lost during compression without sacrificing efficiency.

Table~\ref{tab:qwen25_7b_results} complements the main results with an evaluation on Qwen2.5-7B using the same benchmarks and metrics as Table~\ref{tab:main_results}.

\begin{table}[htbp]
\centering
\caption{Additional results on Qwen2.5-7B. Pass@1 denotes accuracy (\%), and \#L denotes the average number of generated reasoning or latent tokens.}
\label{tab:qwen25_7b_results}
\resizebox{\textwidth}{!}{
\begin{tabular}{l|cc|cc|cc|cc|ccc}
\toprule
& \multicolumn{2}{c|}{GSM} & \multicolumn{2}{c|}{GSM-Hard} & \multicolumn{2}{c|}{SVAMP} & \multicolumn{2}{c|}{MultiArith} & \multicolumn{3}{c}{Average} \\
& Pass@1 & \# L & Pass@1 & \# L & Pass@1 & \# L & Pass@1 & \# L & Pass@1 & \# L & $\Delta_{acc}/L$ \\
\midrule
\rowcolor{blue!5} \multicolumn{12}{c}{\textit{\textbf{Qwen2.5-7B}}} \\
Base CoT & 91.3 & 309 & 54.7 & 393 & 93.4 & 224 & 100 & 195 & 84.85 & 280 & - \\
Tokenskip & 83.8 & 172 & 46.9 & 262 & 90.2 & 124 & 99.40 & 101 & 80.07 & 164 & - \\
\rowcolor[gray]{.9}\textbf{Ours-SFT} & 71.27 & 86.04 & 34.30 & 136.01 & 83.60 & 57.63 & 94.20 & 46.62 & 70.84 & 81.58 & - \\
\rowcolor[gray]{.9}\textbf{Ours-RL} & 76.37 & 84.69 & 39.30 & 135.42 & 87.60 & 55.89 & 99.20 & 45.12 & 75.62 & 80.28 & - \\
\bottomrule
\end{tabular}
}
\end{table}

\section{Limitations}
\label{sec:appendix_limitations}

Despite these promising results, our current framework primarily explores selective compression at the fine-grained token-span level, leaving coarse-grained granularities—such as sentence- or step-level reasoning—largely unexplored. 
In practice, natural reasoning chains and program code exhibit strong structural coherence and semantic modularity. 
Extending selective latent thinking to sentence- or paragraph-level semantic units presents a promising avenue to better capture cohesive logical blocks, which may unlock substantially higher compression ratios while preserving overall reasoning fidelity. 
We leave this structural and hierarchical extension to future work.

\end{document}